\documentclass[screen]{acmart}
\AtBeginDocument{%
  }

\setcopyright{acmlicensed}
\copyrightyear{2018}
\acmYear{2018}
\acmDOI{XXXXXXX.XXXXXXX}

\acmJournal{JACM}
\acmVolume{37}
\acmNumber{4}
\acmArticle{111}
\acmMonth{8}
\usepackage{pznotations}
\usepackage{cleveref}
\usepackage{subcaption} 
\usepackage{nicefrac}
\usepackage{bbding}
\usepackage{bbm}
\usepackage[edges]{forest}

\tikzset{%
	parent/.style = {align=center,text width=0.4cm,rounded corners=3pt, line width=0.3mm, fill=gray!10,draw=gray!80},
	child/.style = {align=center,text width=2.3cm,rounded corners=3pt, fill=blue!10,draw=blue!80,line width=0.3mm},
	grandchild/.style =      {align=center,text width=2cm,rounded corners=3pt},
	greatgrandchild/.style = {align=center,text width=1.5cm,rounded corners=3pt},
	greatgrandchild2/.style = {align=center,text width=1.5cm,rounded corners=3pt},    
	referenceblock/.style =  {align=center,text width=1.5cm,rounded corners=2pt},
	single/.style =           {align=center,text width=3cm,rounded corners=3pt, fill=blue!10,draw=blue!80,line width=0.3mm}, 
	single_work/.style =           {align=center, text width=7cm,rounded corners=3pt, fill=blue!10,draw=blue!0,line width=0.3mm},  
        single_work2/.style =           {align=center, text width=4cm,rounded corners=3pt, fill=blue!10,draw=blue!0,line width=0.3mm},  
	finite/.style =           {align=center,text width=3cm,rounded corners=3pt, fill=red!10,draw=red!80,line width=0.3mm},   
	finite_work/.style =           {align=center,text width=7cm,rounded corners=3pt, fill=red!10,draw=red!0,line width=0.3mm},        
	infinite/.style =           {align=center,text width=3cm,rounded corners=3pt, fill= orange!10,draw= orange!80,line width=0.3mm},   
	infinite_work/.style =           {align=center,text width=7cm,rounded corners=3pt, fill= orange!10,draw= orange!0,line width=0.3mm}, 
}
\begin{document}

%%
%% The "title" command has an optional parameter,
%% allowing the author to define a "short title" to be used in page headers.
\title{Gradient-Based Multi-Objective Deep Learning: Algorithms, Theories, Applications, and Beyond}

%%
%% The "author" command and its associated commands are used to define
%% the authors and their affiliations.
%% Of note is the shared affiliation of the first two authors, and the
%% "authornote" and "authornotemark" commands
%% used to denote shared contribution to the research.
\author{Weiyu Chen}
\authornote{These authors contributed equally to this research.}
%\orcid{1234-5678-9012}
\affiliation{%
  \institution{The Hong Kong University of Science and Technology}
  \city{Hong Kong}
  \country{China}
}
 \email{wchenbx@connect.ust.hk}

 \author{Baijiong Lin}
\authornotemark[1]
\affiliation{%
  \institution{The Hong Kong University of Science and Technology (Guangzhou)}
  \city{Guangzhou}
  \country{China}
}
\email{bj.lin.email@gmail.com}

\author{Xiaoyuan Zhang}
\authornotemark[1]
\authornotemark[2]
\affiliation{%
  \institution{City University of Hong Kong}
  \state{Hong Kong}
  \country{China}}
\email{xzhang2523-c@my.cityu.edu.hk}

\author{Xi Lin}
\affiliation{%
 \institution{City University of Hong Kong}
 \state{Hong Kong}
 \country{China}}
\email{xi.lin@my.cityu.edu.hk}

\author{Han Zhao}
\authornote{Corresponding authors.}
\affiliation{%
\institution{University of Illinois Urbana-Champaign}
\state{Illinois}
\country{USA}}
\email{hanzhao@illinois.edu}

\author{Qingfu Zhang}
\authornotemark[2]
\affiliation{%
  \institution{City University of Hong Kong}
  \state{Hong Kong}
  \country{China}}
\email{qingfu.zhang@cityu.edu.hk}

\author{James T. Kwok}
\authornotemark[2]
\affiliation{%
  \institution{The Hong Kong University of Science and Technology}
  \city{Hong Kong}
  \country{China}}
\email{jamesk@cse.ust.hk}

%%
%% By default, the full list of authors will be used in the page
%% headers. Often, this list is too long, and will overlap
%% other information printed in the page headers. This command allows
%% the author to define a more concise list
%% of authors' names for this purpose.
\renewcommand{\shortauthors}{Chen et al.}
%%
%% The abstract is a short summary of the work to be presented in the
%% article.
\begin{abstract}
Many modern deep learning applications
require balancing multiple
objectives
that are often conflicting.
Examples include multi-task learning, fairness-aware learning, and the alignment of Large Language Models (LLMs).
This leads to multi-objective deep learning, which 
tries to find optimal trade-offs or Pareto-optimal solutions
by adapting mathematical principles from the field of Multi-Objective Optimization (MOO).
However, 
directly applying gradient-based MOO techniques to deep neural networks
presents unique challenges, including high computational costs, optimization instability, and the difficulty of effectively incorporating user preferences. This paper provides a comprehensive survey of gradient-based techniques for multi-objective deep learning. We systematically categorize existing algorithms based on their outputs: (i) methods that find a single, well-balanced solution, (ii) methods that generate a finite set of diverse Pareto-optimal solutions, and (iii) methods that learn a continuous Pareto set of solutions. In addition to this taxonomy, the survey covers theoretical analyses, key applications, practical resources, and highlights open challenges and promising directions for future research. A comprehensive list of multi-objective deep learning algorithms is available at \url{https://github.com/Baijiong-Lin/Awesome-Multi-Objective-Deep-Learning}. 
\end{abstract}

%%
%% The code below is generated by the tool at http://dl.acm.org/ccs.cfm.
%% Please copy and paste the code instead of the example below.
%%

% \ccsdesc[300]{Theory of computation~Mathematical optimization}

%%
%% Keywords. The author(s) should pick words that accurately describe
%% the work being presented. Separate the keywords with commas.

\received{20 February 2007}
\received[revised]{12 March 2009}
\received[accepted]{5 June 2009}

%%
%% This command processes the author and affiliation and title
%% information and builds the first part of the formatted document.
\maketitle

\section{Introduction} \label{sec:intro}

Traditional deep learning often focuses on optimizing a single learning objective, such as minimizing the prediction error or maximizing the likelihood. However, many real-world applications require balancing multiple, often conflicting, objectives. For instance, 
a computer vision system might need to perform the tasks of segmentation, depth estimation, and surface normal prediction simultaneously~\cite{vandenhende2021multi}, thus moving from
single-task learning to multi-task learning~\cite{zhang2021survey}.
Similarly, Large Language Models (LLMs) are expected to excel at diverse tasks like reasoning and coding, while also being safe, fair, and harmless~\cite{wang2023aligning}. 
Multi-Objective Optimization (MOO)~\cite{miettinen1999nonlinear}, a field originating from operations research~\cite{ehrgott2005multicriteria}, provides a formal framework for navigating these trade-offs. It has been widely studied across science and engineering, with diverse applications such as in finance~\cite{steuer2003multiple}, engineering design~\cite{marler2004survey}, and transportation planning~\cite{tzeng2005multi}.

Recently, there has been a surge of interests in adapting MOO for deep learning, re-framing many popular learning problems as multi-objective optimization problems. Examples include multi-task learning where each task performance is considered as an objective~\cite{mtlasmoo}, fairness-aware learning where accuracy is one objective and fairness metrics constitute the other objectives~\cite{martinez2020minimax}, LLM alignment where each alignment criterion (such as helpfulness, harmlessness, and honesty) represents a separate objective~\cite{wang2023aligning}, and federated learning where the performance on each client is treated as an individual objective~\cite{hu2022federated}. All these problems can be formally expressed as:
\begin{align}
	\min_{\vtheta \in \gK \subset \R^d} ~~ \vf(\vtheta) & := [f_1(\vtheta), \dots, f_m(\vtheta)]^\top,
	\label{eq:moo}
\end{align}
where $m \geq 2$ is the number of objectives, $\vtheta$ is the decision variable, $\gK \subset \sR^d$ is the feasible set of the decision variable, and each $f_i: \gK \to \sR$ represents an individual objective function to be minimized.

Unlike single-objective optimization, which focuses on finding a single best solution, MOO recognizes that no single solution is optimal for all objectives simultaneously. Instead, MOO aims to identify a set of solutions that represent different trade-offs between objectives, collectively known as the Pareto set~\cite{miettinen1999nonlinear}. The Pareto set consists of all solutions where improving any objective would necessarily worsen at least one other objective, representing the best possible trade-offs available. In real-world applications, users often have varying preferences for these trade-offs. For instance, in the development of LLMs, a customer service application might emphasize harmlessness and safety, whereas an educational tool might prioritize reasoning and factual accuracy. To address this variability, users can specify their preferences using a vector, $\valpha = [\alpha_1, \dots, \alpha_m]^\top \in \simplex$, where $\simplex = \{\valpha \in \sR^m_+ : \sum_{i=1}^m \alpha_i = 1\}$ is the probability simplex and each $\alpha_i$ represents the importance assigned to the $i$-th objective. MOO approaches can accommodate these user-defined preferences, enabling the discovery of solutions that align with individual needs.

MOO methods can be broadly divided into gradient-free and gradient-based approaches. Gradient-free methods, such as evolutionary algorithms~\cite{deb2002fast,zhang2007moea} and particle swarm optimization~\cite{coello2004handling}, explore the solution space using population-based sampling. While effective for traditional low-dimensional MOO problems~\cite{deb2011multi}, they struggle with deep neural networks, which often have millions or even billions of parameters. Due to the curse of dimensionality, the parameter space becomes too large to explore efficiently without gradient information, leading to slow or failed convergence.  
This makes gradient-based methods~\cite{MGDA, MGDA_0, MGDA_1, CAGrad} preferable for deep neural networks, as they efficiently guide the search using objective gradients.

However, gradient-based MOO faces several significant challenges in deep learning.
Firstly, incorporating user preferences is difficult because directly weighting objectives often produces unaligned solutions.
Secondly, training even one neural network is computationally expensive, posing a major challenge to efficiently approximating the entire Pareto set of networks.
Finally, the use of mini-batch optimization in deep learning introduces high variance, which can further destabilize the optimization process. These challenges have motivated extensive research efforts to develop specialized MOO techniques for deep learning. This paper provides a comprehensive survey of these advances, systematically categorizing existing approaches and identifying key open problems in the field.

\subsection{Comparison with Related Surveys on MOO}
Some surveys have attempted to cover aspects of MOO. For example, Zhang and Yang \cite{zhang2021survey} provide a comprehensive overview of traditional multi-task methods within the machine learning domain. Crawshaw \cite{crawshaw2020multi} offers a summary of multi-task methods in deep learning up to the year 2020. Yu et al. \cite{yu2024unleashing} present a comprehensive survey that covers multi-task learning from traditional machine learning to deep learning. Eichfelder \cite{eichfelder2021twenty} discusses advancements in MOO from a general optimization perspective, without specific emphasis on deep learning. Wei et al. \cite{wei2021review} focus on evolutionary MOO algorithms, but these algorithms are usually not suitable for deep neural networks due to their large parameter spaces and computational demands. Peitz and Hotegni \cite{peitz2024multi} survey the MOO algorithms for deep learning but cover only a limited selection of gradient-based MOO methods and do not include theory and applications. Other surveys concentrate on applications of MOO in areas such as dense prediction tasks \cite{vandenhende2021multi}, recommend systems \cite{zhang2023advances}, and natural language processing \cite{chen2024multi}.

\begin{figure}[!t]
	\centering
	\includegraphics[width=.9\linewidth]{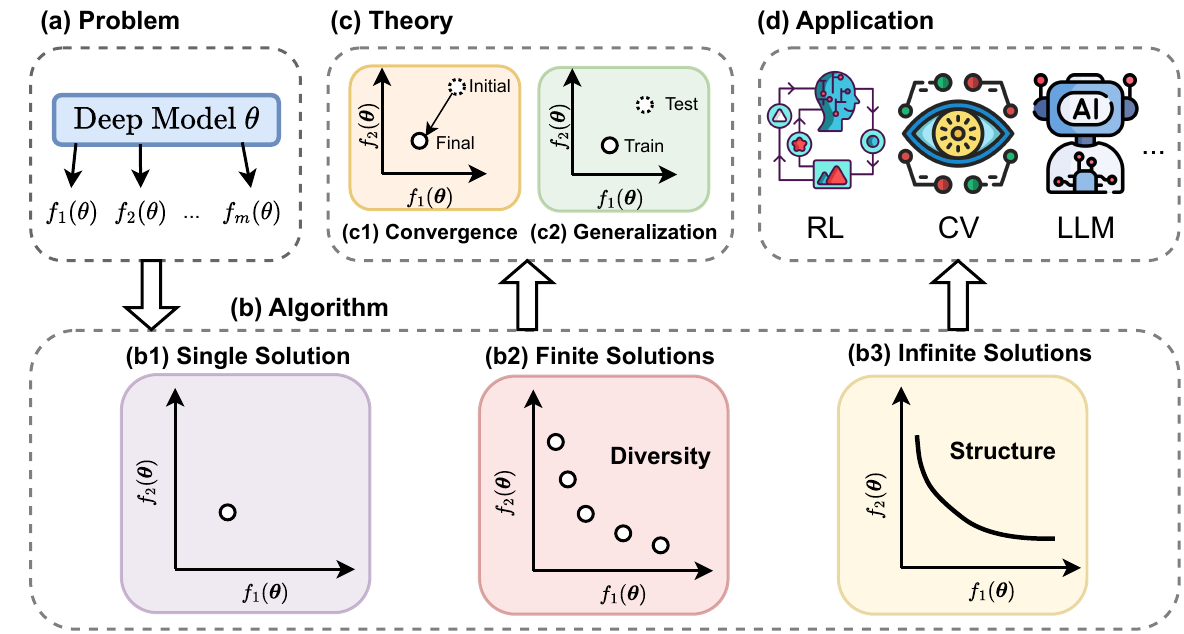}
	\caption{An overview of gradient-based multi-objective deep learning.}
	\label{fig:overview}
\end{figure}

Different with the previous surveys, this survey focuses on recent gradient-based MOO methods in deep learning, encompassing approaches for finding a single solution, a finite set of solutions, and an infinite set of solutions, and covering algorithms, theories, and applications, as the overview shown in Figure \ref{fig:overview}. To the best of our knowledge, it is the first survey paper focusing on the gradient-based MOO methods in deep learning.

\begin{figure*}[!t]
	\footnotesize
	\centering
	\resizebox{\textwidth}{!}{
		\begin{forest}
			for tree={
				forked edges,
				grow'=0,
				draw,
				rounded corners,
				node options={align=center,},
				text width=2.7cm,
				s sep=6pt,
				calign=child edge, calign child=(n_children()+1)/2,
				% l sep = 10 pt,
				% l=40pt
			},
			[\rotatebox{90}{Gradient-based Multi-objective Deep Learning}, fill=gray!45, parent
			% single
			[Finding a Single Solution (\S\ref{sec:single_model}), for tree={single}
			[Loss Balancing Methods (\S\ref{sec:loss_balancing}), single
			[DWA \cite{ljd19}; UW \cite{kendall2018multi}; IMTL-L \cite{liu2021imtl}; Lin et al. \cite{lin2023scale}; MOML \cite{ye2021multi,ye2024moml}; Auto-$\lambda$ \cite{liuauto}; FORUM \cite{ye2024first}; RW \cite{rlw}; STCH \cite{smooth}, single_work] ] [Gradient Balancing Methods (\S\ref{sec:grad_balancing}),  single
			[Gradient Weighting Methods (\S\ref{sec:grad_weighting}),  single
			[MGDA \cite{mtlasmoo}; CAGrad \cite{CAGrad}; SMG \cite{SMGDA}; CR-MOGM \cite{zhou2022convergence}; MoCo \cite{fernando2023mitigating}; PSMGD \cite{xu2024psmgd}; MoDo \cite{chen2024three}; SDMGrad \cite{xiao2023direction}; SGSMGrad \cite{zhang2024convergence}; IMTL-G \cite{liu2021imtl}; Nash-MTL~\cite{NashMTL}; FairGrad~\cite{fairgrad}; UPGrad \cite{quinton2024jacobian}; Aligned-MTL~\cite{alignment}; GradNorm \cite{chen2018gradnorm}; DB-MTL \cite{lin2023scale}; FAMO \cite{famo}, single_work2] 
			]
			[Gradient Manipulation Methods (\S\ref{sec:grad_manipulation}),  single
			[PCGrad \cite{pcgrad}; GradVac~\cite{gradientvacc}; GradDrop \cite{graddrop}, single_work2] 
			]
			]
			]
			% finite
			[Finding a Finite Set of Solutions (\S\ref{sec:pareto_set}), for tree={finite}
			[Methods Based on Preference Vectors (\S\ref{sec:preference_aware}),  finite
			[PMTL~\cite{PMTL}; EPO~\cite{EPO, EPO2}; WC-MGDA~\cite{WCMGDA}; PMGDA~\cite{zhang2024pmgda}; FERERO~\cite{chenferero}; GMOOAR~\cite{gmooar}; UMOD~\cite{zhang2024gliding}, finite_work]
			]
			[Methods without Using Preference Vectors (\S\ref{sec:preference_free}),  finite
			[GradHV~\cite{deist2021multi,deist2020moo,emmerich2007gradient, wang2017hypervolume}; MOO-SVGD~\cite{liu2021profiling}; F4M~\cite{lin2024few}; SoM~\cite{ding2024efficient}; MosT~\cite{li2024many}, finite_work]
			]
			]
			% infinite
			[Finding an Infinite Set of Solutions (\S\ref{sec:pareto_infinite}), for tree={infinite}
			[Hypernetwork-based Methods (\S\ref{sec:hypernetwork}),  infinite
			[PHN \cite{PHN}; CPMTL \cite{cpmtl}; PHN-HVI~\cite{PHNHVI}; SUHNPF \cite{SUHNPF}; Hyper-Trans \cite{tuan2024hyper}, infinite_work]
			]
			[Preference-Conditioned Network-based Methods (\S\ref{sec:pref_net}),  infinite
			[COSMOS \cite{COSMOS}; YOTO \cite{dosovitskiy2019you}; Raychaudhuri et al. \cite{raychaudhuri2022controllable}, infinite_work]
			]
			[Model Combination-based Methods (\S\ref{sec:combine}),  infinite
			[PaMaL \cite{PAMAL}; LORPMAN~\cite{LORPMAN}; Chen and Kwok~\cite{chen2024you}; Panacea~\cite{panacea}; Tang et al.~\cite{tang2024towards}; PaLoRA~\cite{dimitriadis2024pareto}, infinite_work]
			]
			]              
			]
		\end{forest}
	}
	\caption{Taxonomy of existing gradient-based multi-objective deep learning algorithms.}
	\label{fig:taxonomy}
\end{figure*}

\subsection{Organization of the Survey}
In this survey, we categorize gradient-based multi-objective deep learning algorithms based on the type of output they obtain:
(\Cref{fig:taxonomy}): (i) Algorithms that obtain a single, well-balanced model that is not overly biased toward any one objective:
This is achieved through techniques like loss balancing and gradient balancing, which adjust the weight of each objective to resolve conflicts during optimization. (ii) Algorithms that obtain a finite set of Pareto-optimal solutions that allows users to select from multiple options based on their specific needs:
This is accomplished by either dividing the problem into subproblems using preference vectors or directly optimizing for multiple solutions simultaneously. (iii) Algorithms that obtain a continuous set of Pareto-optimal models that can be generated on-demand for any given preference:
This is achieved by learning a solution subspace using efficient structures, where preference vectors are randomly sampled during training and the optimization adapts techniques from single or finite solution methods.

The rest of this survey is organized as follows: Section~\ref{sec:prelim} presents MOO preliminaries, including MOO definitions and concepts; Section \ref{sec:single_model} discusses methods for finding a single Pareto-optimal solution, covering loss balancing and gradient balancing approaches;  \Cref{sec:pareto_set} focuses on methods for identifying a set of finite Pareto-optimal solutions; \Cref{sec:pareto_infinite} covers methods for identifying a set of infinite Pareto optimal solutions; Section~\ref{sec:theory} delves into the theoretical analysis of convergence and generalization in gradient-based multi-objective optimization algorithms; Section \ref{sec:application} showcases various applications in deep learning, including reinforcement learning, Bayesian optimization, computer vision, neural architecture search, recommender systems, and large language models; Section \ref{sec:resource} offers useful resources on datasets and software tools for multi-objective deep learning; Section \ref{sec:future} highlights challenges in the field and suggests promising directions for future research; Section \ref{sec:conclusion} summarizes this survey.

\subsection{Notations}
The notations used in this survey are summarized in \Cref{tab:notation}.
Scalars are represented by non-bold letters; vectors and matrices are denoted by boldface type. Subscripts are used as indices for elements, and superscripts typically denote iteration numbers or indices within a solution set.

\begin{table}[!t]
\caption{Summary of Notations.}
\vspace{-0.1in}
\label{tab:notation}
\centering
% \resizebox{0.6\linewidth}{!}{
\begin{tabular}{ll}
\toprule
Notation & Description \\
\midrule
$\vtheta \in \gK \subset \R^d$ & Decision variable $\vtheta$ in feasible set $\mathcal{K}$ with dimension $d$. \\
$m$ & Number of objectives. \\
$n$ & Number of finite Pareto solutions. \\
$K$ & Number of iterations.\\
$\vf = [f_1,\dots, f_m]^\top$ & Objective function.\\
$\vlam = [\lambda_1, \dots, \lambda_m]^\top$ & Objective weight vector. \\
$\valpha = [\alpha_1, \dots, \alpha_m]^\top$ & Preference vector. \\
$\vz^* = [z_1^*, \dots, z_m^*]^\top$ & Ideal objective value.\\
$\vtheta^{(k)}$ & Solution at $k$-th iteration. \\
$\vtheta^{(1)}, \dots, \vtheta^{(n)}$ & Solution $1$ to Solution $n$ in a size-$n$ solution set. \\
$\vg^{(k)}_i$ & The gradient vector of $i$-th objective at $k$-th iteration. \\
$\mG^{(k)}=[\vg_1^{(k)},\dots,\vg_m^{(k)}]$ & Gradient matrix at $k$-th iteration.  \\ 
$\vd^{(k)}$ & Updating direction of $\vtheta$ at $k$-th iteration. \\
$\vphi$ & Parameters of Pareto set learning structures.\\
$\mathrm{HV}_\vr(\cdot)$ & Hypervolume with respect to reference point $\vr$. \\
$[m]$ & Index set $\{1, \dots, m \}$.\\
$\simplex$ & $(m-1)$-D simplex $\{\valpha | \sum_{i=1}^m \alpha_i=1, \alpha_i \geq 0, i\in[m] \}$. \\
$\|\cdot\|$ & $\ell_2$ norm. \\
$\eta$ & Step size for updating $\vtheta$. \\
$\epsilon$ & Error tolerance. \\
\bottomrule
\end{tabular}%}
\end{table}

\section{Preliminary: Multi-Objective Optimization} \label{sec:prelim}

In this section, we first introduce some MOO concepts in \Cref{sec:pareto_concept} and then review two classical methods (i.e., linear scalarization and Tchebycheff scalarization) to find Pareto solutions of MOO problems in defined in Section \ref{sec:scalarization}. 

\subsection{Key Concepts in Multi-Objective Optimization} \label{sec:pareto_concept}

Unlike single-objective optimization, solutions in MOO 
cannot be directly compared based on a single criterion, but are compared using the concept of dominance as follows. Note that without loss of generality, we consider multi-objective minimization (i.e., problem (\ref{eq:moo})) here.
\begin{definition}[(strict) Pareto dominance~\cite{miettinen1999nonlinear}]
	A solution $\vtheta^{(a)}$ dominates another solution $\vtheta^{(b)}$ (denoted as $\vtheta^{(a)} \preceq \vtheta^{(b)}$) if and only if $f_i(\vtheta^{(a)}) \leq f_i(\vtheta^{(b)})$ for all $i \in [m]$, and there exists at least one $i \in [m]$ such that $f_i(\vtheta^{(a)}) < f_i(\vtheta^{(b)})$.
	A solution $\vtheta^{(a)}$ \textit{strictly} dominates another solution $\vtheta^{(b)}$ if and only if $f_i(\vtheta^{(a)}) < f_i(\vtheta^{(b)})$ for all $i \in [m]$.
\end{definition}
Based on this definition, we further define Pareto optimality (PO), Pareto set, and Pareto front as follows.

\begin{definition}[(weak) Pareto optimality~\cite{miettinen1999nonlinear}] \label{def:po}
	A solution $\vtheta^*$ is Pareto optimal if no other solution dominates it. A solution $\vtheta^*$ is \textit{weakly} Pareto optimal if no other solution \textit{strictly} dominates it.
\end{definition}

\begin{definition}[Pareto set (PS) and Pareto front (PF)~\cite{miettinen1999nonlinear}] \label{def:ps_pf}
	A PS is the set of all PO solutions. A PF is the set of all objective function values of the PO solutions.
\end{definition}

\begin{figure}[!t]
\centering
\includegraphics[width=0.30\linewidth]{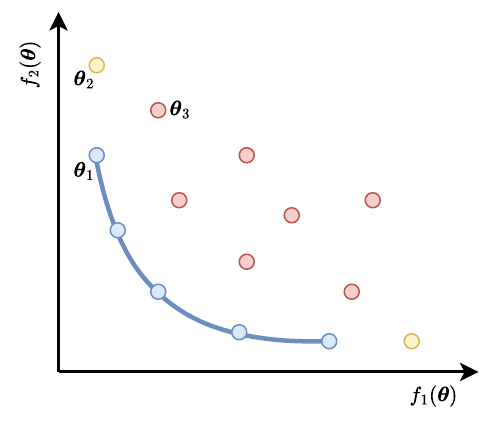}
\vspace{-0.1in}
\caption{Illustration of Pareto optimality concepts. The blue, yellow, and red circles denote Pareto optimal solutions, weakly Pareto optimal solutions, and dominated solutions, respectively. The blue curve represents the Pareto front.}
\label{fig:pareto}
\end{figure}

Figure~\ref{fig:pareto} illustrates the Pareto concepts on the two-objective problem. The blue circles represent \textbf{Pareto optimal solutions} (defined in Definition \ref{def:po}), which indicates that no solution can improve one objective without worsening the other. The blue curve connecting the blue circles denotes the \textbf{Pareto front}, as defined in Definition \ref{def:ps_pf}. The yellow circles indicate \textbf{weak Pareto optimal solutions} since they can be improved in one objective without negatively impacting the other. For example, comparing $\vtheta_2$ and $\vtheta_1$, $\vtheta_2$ is a weak Pareto optimal solution since it can be improved in the second objective without affecting the first. The red circles represent solutions that are \textbf{not Pareto optimal} because there exists another solution that strictly Pareto dominates them. For instance, comparing $\vtheta_3$ and $\vtheta_1$, $\vtheta_3$ is not a Pareto optimal solution as $\vtheta_1$ outperforms it in both objectives.

\begin{definition}[Preference vector]
	\label{def:pv}
	$\valp$ denotes a preference vector. The value of the entries of $\valp$ denote the preference of a specific objective. Throughout this paper, a preference vector $\valp$ is constrained on a simplex $\simplex$, where $\simplex = \{\valpha | \sum_{i=1}^m \alpha_i=1, \alpha_i \geq 0, i\in[m] \}$.
\end{definition}

In deep learning, the parameter $\vtheta$ is typically optimized using gradient descent. In single-objective optimization, when the objective function is non-convex, the optimization process often reaches a stationary point. In the case of MOO, the solution generally converges to a \emph{Pareto stationary} solution, which is defined as follows:

\begin{definition}[Pareto stationary~\cite{MGDA}]
	\label{def:stationary}
	A solution $\vtheta^*$ is called \emph{Pareto stationary} if there exists a vector $\vlam \in \simplex$ such that $\| \sum_{i=1}^m \lambda_i \nabla f_i(\vtheta^*) \| = 0$.
\end{definition}
Pareto stationarity is a necessary condition for achieving Pareto optimality. If all objectives are convex with $\lambda_i > 0 , \forall i$, it also serves as the Karush-Kuhn-Tucker (KKT) sufficient and necessary condition for Pareto optimality~\cite{MGDA}.  
When evaluating the performance of a set of obtained solutions, the Hypervolume (HV)~\cite{zitzler1998multiobjective} indicator is one of the most widely used performance indicators, which is defined as follows.

\begin{definition}[Hypervolume~\cite{zitzler1998multiobjective}] \label{def:hv}
	Given a solution set $\sS=\{\vq^{(1)},\dots,\vq^{(N)}\}$ and a reference point $\vr$, the hypervolume of $\sS$ is calculated by:
	\begin{align}
		\mathrm{HV}_\vr (\sS) = \Lambda({\vp\;|\;\exists \vq \in \sS: \vq \preceq \vp \preceq \vr }),
	\end{align}
	\noindent where $\Lambda (\cdot)$ denotes the Lebesgue measure of a set. 
\end{definition}
HV provides a unary measure of both convergence and diversity of a solution set. A larger HV indicates that the obtained solutions are more diverse and closer the PF in the objective space.

\subsection{Scalarization Methods} \label{sec:scalarization}
\textbf{Linear Scalarization (LS).} The most straightforward way to solve MOO problems is to reformulate them as single-objective optimization problems weighted by the given preference vector $\valp$ as follows: 
\begin{equation}
	\label{eqn:ls}
	\min_{\vtheta \in \gK} g_\valp^\mathrm{LS}(\vtheta) := \sum_{i=1}^m \alpha_i f_i(\vtheta).    
\end{equation}
LS is widely used in practice due to its simplicity. However, this method suffers from a limitation: it can only identify solutions on the convex part of the PF. For PFs with a concave shape, using LS only finds the endpoints of a PF. 

\vspace{0.1in}
\noindent \textbf{Tchebycheff Scalarization.} Another way to convert a MOO problem to a single objective optimization problem is to use the Tchebycheff scalarization function~\cite{tch1,tch2}, defined as:
\begin{equation}
	\label{eq:tch}
	\min_{\vtheta \in \gK} g_\valp^\mathrm{Tche}(\vtheta) := \max_{i \in [m]} \alpha_i (f_i(\vtheta) - z_i^*),
\end{equation}
where \(z_i^*\) are reference values, usually set as the ideal value of the \(i\)-th objective. Unlike linear scalarization, Tchebycheff scalarization can explore the entire PF by traversing all preference vectors within the simplex. This ensures that both convex and concave regions of the PF are covered. The relationship between Tchebycheff scalarization and weak Pareto optimality can be formally described as follows:

\begin{theorem}[\cite{choo1983proper}]
	\label{prop:tch}
	A solution \(\vtheta^*\) 
	of the original MOO problem (\ref{eq:moo})
	is weakly Pareto optimal 
	if and only if there exists a preference vector \(\valpha\) such that \(\vtheta^*\) is an optimal solution of problem (\ref{eq:tch}).
\end{theorem}

If $ \vtheta^* $ is unique for a given $\valp$, it is considered Pareto optimal. However, Tchebycheff scalarization poses challenges for gradient-based deep learning due to the nonsmooth nature of the $ \max(\cdot) $ operator. This nonsmoothness leads to non-differentiability and a slow convergence rate of $ \mathcal{O}(1/\epsilon^2)$, where $ \epsilon $ represents the error tolerance~\cite{smooth}. Even when all objectives $ \{f_i\}_{i=1}^m $ are differentiable, problem~(\ref{eq:tch}) is still non-differentiable.

\section{Finding a Single Pareto Optimal Solution} \label{sec:single_model}
\begin{figure*} 
	\centering
	\subfloat[EW.]{%
		\includegraphics[width=0.23\linewidth]{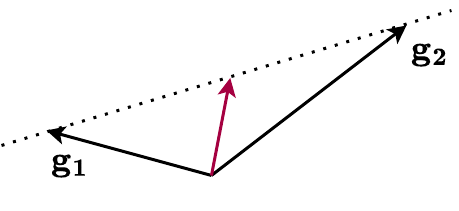}}
	\hfill
	\subfloat[DWA \cite{ljd19}.]{%
		\includegraphics[width=0.23\linewidth]{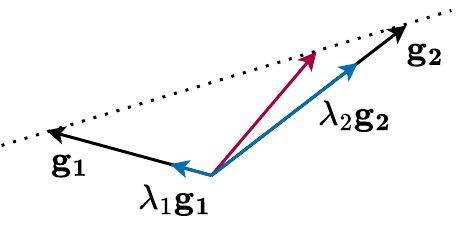}}
	\hfill
	\subfloat[MGDA \cite{mtlasmoo}.]{%
		\includegraphics[width=0.23\linewidth]{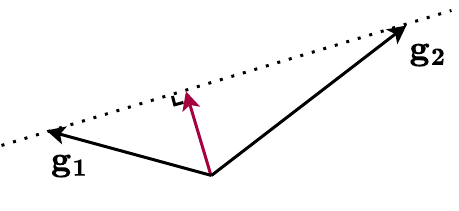}}
	\hfill
	\subfloat[PCGrad \cite{pcgrad}.]{%
		\includegraphics[width=0.23\linewidth]{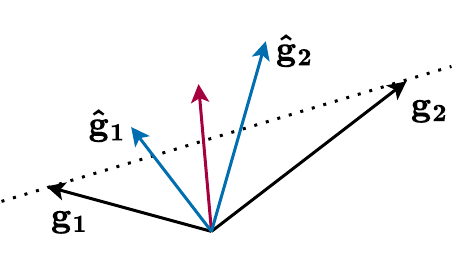}}
	\caption{Illustration of the update direction $\vd$ of several representative methods: EW, DWA \cite{ljd19} from loss balancing methods, MGDA \cite{mtlasmoo} from gradient weighting methods, and PCGrad \cite{pcgrad} from gradient manipulation methods. They are illustrated in a two-objective learning problem (two specific gradients are labeled as $\vg_1$ and $\vg_2$).}
	\label{fig:mtl_direction} 
\end{figure*}

In some scenarios such as multi-task learning (MTL)~\cite{zhang2021survey,ruder2017overview,crawshaw2020multi}, it is sufficient to identify a single Pareto optimal solution without the need to explore the entire Pareto set. 
Here, all objectives are usually treated as equally important (i.e., the preference vector is $\valp=[\frac{1}{m}, \dots, \frac{1}{m}]^\top$), and the goal is to produce a single model that effectively balances all objectives. 
The most straightforward method is linear scalarization (i.e., problem (\ref{eqn:ls})) with $\alpha_i=\frac{1}{m}$, which is known as Equal Weighting (EW) \cite{zhang2021survey} in MTL. However, EW may cause some tasks to have unsatisfactory performance~\cite{standley2020tasks, lin2023scale}. Therefore, to improve the performance, many methods have been proposed to dynamically tune the objective weights $\{\lambda_i\}_{i=1}^m$ during training, which can in general be formulated as:
\begin{equation}
	\min_{\vtheta} \sum_{i=1}^m \lambda_i f_i(\vtheta),
	\label{eq:mtl}
\end{equation}
where $\lambda_i$ is the weight for the $i$-th objective $f_i(\vtheta)$ and the uniform preference vector $\valp$ is omitted for simplicity. These methods can be categorized as loss balancing methods and gradient balancing approaches. Some representative loss/gradient balancing methods are illustrated in Figure \ref{fig:mtl_direction} (adapted from \cite{CAGrad, liu2021imtl}).

\subsection{Loss Balancing Methods} \label{sec:loss_balancing}
This type of approach dynamically computes or learns the objective weights $\{\lambda_i\}_{i=1}^m$ during training using some measures on loss such as the decrease speed of loss \cite{ljd19}, homeostatic uncertainty of loss \cite{kendall2018multi}, loss scale~\cite{liu2021imtl, lin2023scale}, and validation loss~\cite{ye2021multi, ye2024moml, ye2024first, liuauto}.

Dynamic Weight Average (DWA) \cite{ljd19} estimates each objective weight as the ratio of the training losses in the last two iterations. Formally, the objective weight in $k$-th iteration ($k>2$) is computed as follows:
\begin{equation}
	\lambda_{i}^{(k)} = \frac{m\exp(\omega^{(k-1)}_i / \gamma)}{\sum_{j=1}^m\exp(\omega_j^{(k-1)}/\gamma)}, \quad \omega^{(k-1)}_j=\frac{f^{(k-1)}_j}{f^{(k-2)}_j},
\end{equation}
where $\gamma$ is the temperature and $f^{(k)}_i$ is the loss value of $i$-th objective at $k$-th iteration. 

Kendall et al.~\cite{kendall2018multi} propose Uncertainty Weighting (UW) that considers the homoscedastic uncertainty of each objective and optimizes objective weights together with the model parameter as follows:
\begin{equation}
	\min_{\vtheta, \vs} \sum_{i=1}^m \left(\frac{1}{2s_i^2}f_i(\vtheta) + \log s_i\right),
\end{equation}
where $\vs=[s_1,\dots,s_m]^\top$ is learnable and $\log s_i$'s are regularization terms.

Impartial Multi-Task Learning (IMTL) \cite{liu2021imtl} balances
losses across tasks, abbreviated
as IMTL-L(oss). Specifically, IMTL-L encourages all objectives to have a similar loss scale by transforming each objective $f_i(\vtheta)$ as $e^{s_i}f_i(\vtheta) - s_i$. Similar to UW \cite{kendall2018multi}, IMTL-L simultaneously learns the transformation parameter $\vs$ and the model parameter $\vtheta$ as follows:
\begin{equation}
	\min_{\vtheta, \vs} \sum_{i=1}^m \left(e^{s_i}f_i(\vtheta) - s_i\right).
\end{equation}

Similar to IMTL-L \cite{liu2021imtl}, Lin et al. \cite{lin2023scale} also aim to make all objective losses have a similar scale. They achieve it by performing a logarithm transformation on each objective (i.e., $\log f_i(\vtheta)$). Moreover, they theoretically show that the transformation in IMTL-L is equivalent to the logarithm transformation when $s_i$ is the exact minimizer of $\min_{s_i} e^{s_i}f_i(\vtheta) - s_i$ in each iteration. Hence, the logarithm transformation can recover the transformation in IMTL-L.

Ye et al. \cite{ye2021multi, ye2024moml} propose Multi-Objective Meta Learning (MOML), which adaptively tunes the objective weights $\vlam$ based on the validation performance, reformulating problem (\ref{eq:mtl}) as a multi-objective bi-level optimization problem:
% \vspace{-0.1in}
\begin{subequations}
	\label{eq:moml}
	\begin{align}
		&\min_{\vlam}~\big[f_1(\vtheta^*(\vlam);\mathcal{D}^{\text{val}}_1),\dots,f_m(\vtheta^*(\vlam);\mathcal{D}^{\text{val}}_m)\big]^\top \label{eq:moml-ul} \\
		&\ ~ \mathrm{s.t.}\ ~ \vtheta^*(\vlam)=\argmin_{{\vtheta}} \sum_{i=1}^m \lambda_i f_i(\vtheta;\mathcal{D}^{\text{tr}}_i),
		\label{eq:moml-ll}
	\end{align}
\end{subequations}
where $\mathcal{D}^{\text{tr}}_i$ and $\mathcal{D}^{\text{val}}_i$ are the training and validation datasets for
the $i$-th objective, respectively. In each training iteration, when given objective weights $\{\lambda_i\}_{i=1}^m$, MOML first learns the model $\vtheta^*(\vlam)$ on the training dataset by solving lower-level (LL) subproblem (\ref{eq:moml-ll}) with $T$ iterations and then updates $\vlam$ in the upper-level (UL) subproblem (\ref{eq:moml-ul}) via minimizing the loss of the trained MTL model $\vtheta^*(\vlam)$ on the validation dataset of each objective. However, when solving the UL subproblem (\ref{eq:moml-ul}), MOML needs to calculate the complex hypergradient $\nabla_{\vlam} \vtheta^*(\vlam)$ which requires to compute many Hessian-vector products via the chain rule. Hence, the time and memory costs of MOML grow significantly fast with respect to the dimension of $\vtheta$ and the number of LL iterations $T$.

Auto-$\lambda$ \cite{liuauto} modifies problem (\ref{eq:moml}) by replacing the multi-objective UL subproblem (\ref{eq:moml-ul}) with a single-objective problem: $\min_{\vlam} \sum_{i=1}^m f_i(\vtheta^*(\vlam); \mathcal{D}^{\text{val}}_i)$.
Besides, Auto-$\lambda$ approximates the hypergradient $\nabla_{\vlam} \vtheta^*(\vlam)$ via the finite difference approach. Thus, it is more efficient than MOML \cite{ye2021multi, ye2024moml}.

FORUM~\cite{ye2024first} is proposed to improve the efficiency of MOML \cite{ye2021multi, ye2024moml}. FORUM first reformulates problem (\ref{eq:moml}) as a constrained MOO problem via the value-function approach \cite{liu2021value} and then proposes a multi-gradient aggregation method to solve it.
Compared with MOML, FORUM is a first-order method and does not need to compute the hypergradient $\nabla_{\vlam} \vtheta^*(\vlam)$. Thus, it is significantly more efficient than MOML.

Random Weighting (RW)~\cite{rlw} randomly samples objective weights from a distribution (such as the standard normal distribution) and normalizes them into the simplex in each iteration. RW has a higher probability of escaping local minima than EW due to the randomness from objective weights, resulting in better generalization performance.

Different from the above methods based on linear scalarization, Lin et al. \cite{smooth} propose Smooth Tchebycheff scalarization (STCH), which transforms the Tchebycheff scalarization into a smooth function: 
\begin{align}
	\label{eq:stch}
	\min_{\vtheta} f_\valp^\mathrm{STCH}(\vtheta, \mu) 
	:= \mu \log \sum_{i=1}^m \exp \sbr{\frac{\alpha_i (f_i(\vtheta) - z_i^*)}{\mu}},
\end{align}
where $\mu > 0$ is a smoothing parameter, and \(\vz^*\) is a reference point as in Tchebycheff scalarization (problem (\ref{eq:tch})).
STCH improves upon the original Tchebycheff scalarization by making \( f_\valp^\mathrm{STCH}(\vtheta, \mu) \) smooth when all objective functions \( f_i(\vtheta) \)'s are smooth, resulting in faster convergence. It also retains the advantages of the original Tchebycheff scalarization: (1) convexity when all objective functions are convex and (2) Pareto optimality by appropriately choosing \( \mu \).

\subsection{Gradient Balancing Methods}\label{sec:grad_balancing}

Let $\vg_i^{(k)}=\nabla_\vtheta f_i (\vtheta)|_{\vtheta^{(k)}}\in \mathbb{R}^d$ be the gradient of the $i$-th objective at the $k$-th iteration.
This type of approach aims to find a common update direction $\vd^{(k)}$ by adaptively aggregating the gradients of all objectives $\{\vg_i^{(k)}\}_{i=1}^m$ at each iteration. Then, the model parameter is updated via $\vtheta^{(k+1)} = \vtheta^{(k)} - \eta \vd^{(k)}$, where $\eta$ is a step size. To simplify notations, we omit the superscript in this section. Compared with loss balancing methods in \Cref{sec:loss_balancing}, gradient balancing approaches usually achieve better performance.

\subsubsection{Gradient Weighting Methods} \label{sec:grad_weighting}

In most gradient weighting methods, $\vd$ is computed by:
\begin{align}
	\label{eq:wsg}
	\vd = \mG \vlam,
\end{align}
where $\mG = [\vg_1, \dots, \vg_m]\in \mathbb{R}^{d \times m}$ is the gradient matrix and $\vlam = [\lambda_1, \dots, \lambda_m]^\top$ is the gradient weights.

Sener and Koltun~\cite{mtlasmoo} apply Multiple-Gradient Descent Algorithm (MGDA) \cite{MGDA}, which aims to find a direction $\vd$ at each iteration to maximize the minimal decrease across the losses. This is formulated as:
\begin{align}
	\max_{\vd} \min_{i \in [m]}(f_i(\vtheta)-f_i(\vtheta - \eta \vd)) \approx \max_{\vd} \min_{i \in [m] } \vg^ \top_i \vd,
	\label{eq:mgda_org}
\end{align}
where $\eta$ is the step size. The solution of problem (\ref{eq:mgda_org}) is $\vd = \mG \vlam$, where $\vlam$ is computed as:
\begin{align}
	\vlam = \argmin_{\vlam\in \simplex}\left\|\mG\vlam\right\|^2, \label{eq:mgda}
\end{align}
where $\simplex$ denotes the simplex. Sener and Koltun~\cite{mtlasmoo} use the Frank-Wolfe algorithm~\cite{jaggi2013revisiting} to solve problem (\ref{eq:mgda}).

Conflict-Averse Gradient descent (CAGrad)~\cite{CAGrad} improves MGDA~\cite{mtlasmoo} by constraining the aggregated gradient $\vd$ to be around the average gradient $\vg_0=\frac{1}{m} \sum_{i=1}^m \vg_i$ with a distance $c\norm{\vg_0}$, where $c \in [0,1)$ is a constant. Specifically, $\vd$ is computed by solving the following problem:
\begin{align} \label{eq:cagrad_org}
	\max_{\vd} \min_{i \in [m]} \vg_i^\top \vd, 
	\quad \mathrm{s.t.} ~~ \norm{\vd-\vg_0} \le c\norm{\vg_0}.
\end{align}
Problem (\ref{eq:cagrad_org}) is equivalent to first computing $\vlam$ by solving the following problem:
\begin{align} \label{eq:cagrad}
	\vlam = \argmin_{\vlam \in \simplex} {\vg^\top_\vlam} \vg_0 + \norm{\vg_0} \norm{\vg_{\vlam}},
\end{align}
where $\vg_\vlam = \frac{1} {m}\mG\vlam$ and then calculate the update direction $\vd = \vg_0 + \frac{c}{\norm{\vg_\vlam}}\vg_\vlam$. Note that CAGrad is simplified to EW when $c=0$ and degenerates into MGDA~\cite{mtlasmoo} when $c\rightarrow +\infty$.

MGDA \cite{mtlasmoo} and its variant CAGrad \cite{CAGrad} can converge to a Pareto stationary point in the deterministic setting with full-gradient computations. However, in deep learning, where stochastic gradients are used, many works \cite{SMGDA,zhou2022convergence,xu2024psmgd,fernando2023mitigating,chen2024three,xiao2023direction,zhang2024convergence} explore the convergence properties of MGDA under stochastic gradients, which are introduced in detail in Section \ref{sec:convergence}.

IMTL-G(rad) \cite{liu2021imtl} aims to find $\vd$ that has equal projections on all objective gradients. Thus, we have:
\begin{align}
	\vu^\top_1 \vd = \vu_i^ \top \vd, \quad 2\le i \le m,
	\label{eq:imtl-g}
\end{align}
where $\vu_i = \nicefrac{\vg_i}{\norm{\vg_i}}$. If constraining $\sum_{i=1}^m \lambda_i=1$, problem (\ref{eq:imtl-g}) has a closed-form solution of $\vlam$:
\begin{align}
	\vlam_{(2,\dots,m)} = \vg_1^\top \mU \left(\mD{\mU}^\top\right)^{-1}, \quad \lambda_1 = 1 - \sum_{i=2}^m \lambda_i,
\end{align}
where $\vlam_{(2,\dots,m)} = [\lambda_2, \dots, \lambda_m]^\top$, $\mU = [\vu_{1} - \vu_2, \dots, \vu_1 - \vu_m ]$, and $\mD = [\vg_1 - \vg_2, \dots, \vg_1 - \vg_m ]$.

Nash-MTL~\cite{NashMTL} formulates problem (\ref{eq:wsg}) as a bargaining game~\cite{nash,bargaining}, where each objective is cast as a player and the utility function for each player is defined as $\vg_i^\top \vd$. Hence, $\vd$ is computed by solving the following problem: 
\begin{align}
	\max_{\vd} \sum_{i=1}^m \log\left({\vg_i}^\top \vd\right).
	\label{eq:nash_org}
\end{align}
The solution of problem (\ref{eq:nash_org}) is $\vd = \mG \vlam$, where $\vlam$ is obtained by solving the following problem:
\begin{align}
	\mG^{\top} \mG \vlam = \vlam^{-1},
	\label{eq:nash}
\end{align}
which can be approximately solved using a sequence of convex optimization problems.

FairGrad~\cite{fairgrad} extends Nash-MTL~\cite{NashMTL} by computing $\vlam$ through the equation $\mG^{\top} \mG \vlam = \vlam^{-1/\gamma}$, where $\gamma \ge 0$ is a constant.
Unlike Nash-MTL \cite{NashMTL}, FairGrad treats the problem as a constrained nonlinear least squares problem.
Similarly, UPGrad \cite{quinton2024jacobian} calculates $\vlam$ by optimizing $\min_{\vlam}\vlam^\top (\mG^\top \mG) \vlam$, a quadratic programming (QP) problem.

Aligned-MTL~\cite{alignment} considers problem (\ref{eq:wsg}) as a linear system and aims to enhance its stability by minimizing the condition number of $\mG$. Specifically, the Gram matrix $\mC = \mG^{ \top} \mG$ is first computed. Then, the eigenvalues $\{\sigma_1, \dots, \sigma_R\}$ and eigenvectors $\mV$ of $\mC$ are obtained. Finally, $\vlam$ is computed by:
\begin{align}
	\vlam = \sqrt{\sigma_R} \mV{\mSigma}^{-1} {\mV}^{\top},
\end{align}
where ${\mSigma}^{-1} = \text{diag}\left(\sqrt{\nicefrac{1}{\sigma_1}}, \dots, \sqrt{\nicefrac{1}{\sigma_R}}\right)$.

Gradient Normalization (GradNorm)~\cite{chen2018gradnorm} aims to learn $\vlam$ so that the scaled gradients have similar magnitudes. It updates $\vlam$ by solving the following problem via one-step gradient descent:
\begin{align}
	\min_{\vlam} \sum_{i=1}^m \left( \norm{\lambda_i \vg_i} - c \times r_i^\gamma \right),
	\label{eq:gradnorm}
\end{align}
where $c=\frac{1}{m} \sum_{i=1}^m  \norm{\lambda_i \vg_i}$ is the average scaled gradient norm and is treated as a constant, $r_i = \frac{f_i/f^{(0)}_i}{\frac{1}{m}\sum_{j=1}^m (f_j / f_j^{(0)})}$ is the training speed of the $i$-th objective and is also considered as a constant, and $\gamma>0$ is a hyperparameter.

Dual-Balancing Multi-Task Learning (DB-MTL)~\cite{lin2023scale} also normalizes all objective gradients to the same magnitude. $\vlam$ is computed as $\lambda_{i} = \nicefrac{\gamma}{\norm{\vg_{i}}}$, where $\gamma = \max_{i \in [m]} \norm{\vg_i}$ is a scaling factor controlling the update magnitude. Thus, compared with GradNorm~\cite{chen2018gradnorm}, this method guarantees all objective gradients have the same norm in each iteration. Moreover, they observe that the choice of the update magnitude $\gamma$ significantly affects performance.

\subsubsection{Gradient Manipulation Methods} \label{sec:grad_manipulation}

To address gradient conflicting ,
gradient manipulation methods correct each objective gradient $\vg_i$ to $\hat{\vg}_i$ 
and then compute the update direction as $\vd=\sum_{i=1}^m \hat{\vg}_i$.

Projecting Conflicting Gradients (PCGrad)~\cite{pcgrad} considers two objective gradients $\vg_i$ and $\vg_j$ as conflicting if
$\vg_i^\top \vg_j < 0$. Then, it corrects each objective gradient by projecting it onto the normal plane of other objectives' gradients. Specifically, at each iteration, $\hat{\vg}_i$ is initialized with its original gradient $\vg_i$. Then, for each $j \neq i$, if
$\hat{\vg}_i^\top \vg_j < 0$, $\hat{\vg}_i$ is corrected as:
\begin{align}
	\label{eq:pcgrad}
	\hat{\vg}_i = \hat{\vg}_i - \frac{\hat{\vg}_i^\top \vg_j}{\norm{\vg_j}^2} \vg_j.
\end{align}
This reduces gradient conflicts by removing the components of $\hat{\vg}_i$ that oppose other objective gradients.

While PCGrad \cite{pcgrad} corrects the gradient if and only if two objectives have a negative gradient similarity, Gradient Vaccine (GradVac)~\cite{gradientvacc} extends it to a more general and adaptive formulation. In each iteration, if $\phi_{ij}<\bar{\phi}_{ij}$, GradVac updates the corrected gradient $\hat{\vg}_i$ as:
\begin{equation}
	\label{eq:gradient_vaccine}
	\hat{\vg}_i = \hat{\vg}_i - \frac{ \norm{\hat{\vg}_i} \left( \bar{\phi}_{ij} \sqrt{1 - \phi_{ij}^2} - \phi_{ij} \sqrt{1 - \bar{\phi}_{ij}^2} \right) }{ \norm{\vg_j} \sqrt{1 - \bar{\phi}_{ij}^2} } \vg_j,
\end{equation}
where $\phi_{ij}$ is the cosine similarity between $\hat{\vg}_i$ and $\vg_j$. $\bar{\phi}_{ij}$ is initialized to $0$ and updated by Exponential Moving Average (EMA), i.e., $\bar{\phi}_{ij}\leftarrow (1-\beta)\bar{\phi}_{ij} + \beta \phi_{ij}$, where $\beta$ is a hyperparameter. Note that GradVac simplifies to PCGrad when $\bar{\phi}_{ij}=0$.
Gradient Sign Dropout (GradDrop) \cite{graddrop} considers that conflicts arise from differences in the sign of gradient values. Thus, a probabilistic masking procedure is proposed to preserve gradients with consistent signs during each update.

\subsubsection{Practical Speedup Strategy}
\label{sec:speedup}
Gradient balancing methods suffer from high computational and storage costs.
Specifically, in each iteration, almost all gradient balancing methods require performing $m$ back-propagation processes to obtain all objective gradients w.r.t. the model parameter $\vtheta \in \mathbb{R}^d$ and then store these gradients $\mG \in \R^{d \times m}$. This can be computationally expensive when dealing with a large number of objectives or using a neural network with a large number of parameters. Moreover, many gradient balancing methods (such as MGDA \cite{mtlasmoo}, CAGrad \cite{CAGrad}, Nash-MTL \cite{NashMTL}, and FairGrad \cite{fairgrad}) need to solve an optimization problem
to obtain the objective weight $\vlam$ in each iteration, which also increases the computational and memory costs. Hence, several strategies are proposed to alleviate this problem.

Sener and Koltun \cite{mtlasmoo} use feature-level gradients (i.e., gradients w.r.t. the representations from the last shared layer) to replace the parameter-level gradients (i.e., gradients w.r.t. the shared parameters $\vtheta$). Since the dimension of the representation is much smaller than that of the shared parameters, it can significantly reduce the computational and memory costs. This strategy is also adopted by some gradient balancing methods, such as IMTL-G \cite{liu2021imtl} and Aligned-MTL \cite{alignment}. 

Liu et al. \cite{CAGrad} randomly select a subset of objectives to calculate the update direction in each iteration.
Navon et al. \cite{NashMTL} propose to update the objective weight $\vlam$ every $\tau$ iterations instead of updating in each iteration. Although this strategy speeds up training, they observe that it may cause a noticeable drop in performance.
Liu et al.~\cite{famo} consider optimizing the logarithm of the MGDA objective \cite{mtlasmoo} and propose a speedup strategy. Specifically, when solving problem (\ref{eq:mgda}) via one-step gradient descent, $\vlam$ is updated as $\vlam\leftarrow \vlam - \eta \nabla_{\vlam}\left\|\mG\vlam\right\|^2$. Note that
\begin{align}
	&\frac{1}{2} \nabla_{\vlam} \norm{\mG\vlam}^2 = {\mG}^\top \mG \vlam = {\mG}^\top \vd  \approx \frac{1}{\eta} \left[f^{(k)}_1 - f^{(k+1)}_1, \dots, f^{(k)}_m - f^{(k+1)}_m\right] ^\top,
\end{align}
where $f^{(k)}_i$ is the loss value of the $i$-th objective in the $k$-th iteration. Hence, $\vlam$ can be approximately updated using the change in losses without computing all objective gradients.
Although this strategy significantly reduces computational and memory costs, it may cause performance degradation.
Moreover, it is only applicable to MGDA-based methods.

\subsection{Discussions}
Loss balancing and gradient balancing methods effectively mitigate conflicts among different objectives and achieve a Pareto-optimal solution, which is important for
MOO scenarios such as multi-task learning. However, they are limited to finding a single Pareto-optimal solution and lack the ability to control the solution's position on the Pareto front, resulting in inapplicability in some scenarios such as multi-criteria learning.
In \Cref{sec:pareto_set,sec:pareto_infinite}, we introduce methods to achieve more diverse and targeted solutions.

The computational cost of loss balancing methods (except MOML~\cite{ye2021multi, ye2024moml}, Auto-$\lambda$~\cite{liuauto}, and FORUM~\cite{ye2024first}) is almost the same as that of EW since they only need one backpropagation process in each training iteration. However, gradient balancing methods are much more computationally expensive due to the need of $m$ backpropagation processes to obtain each objective gradient, where $m$ is the number of objectives.
While various speedup strategies can address this problem, they often come at the cost of performance, as detailed in \Cref{sec:speedup}.
Compared to loss balancing methods, gradient balancing methods generally exhibit better convergence properties such as guaranteeing to converge to a Pareto stationary point, detailed in~\Cref{sec:convergence}.

\begin{figure*}[!t] 
\centering
\subfloat[MGDA \cite{mtlasmoo}.]{
\includegraphics[width=0.18\linewidth]{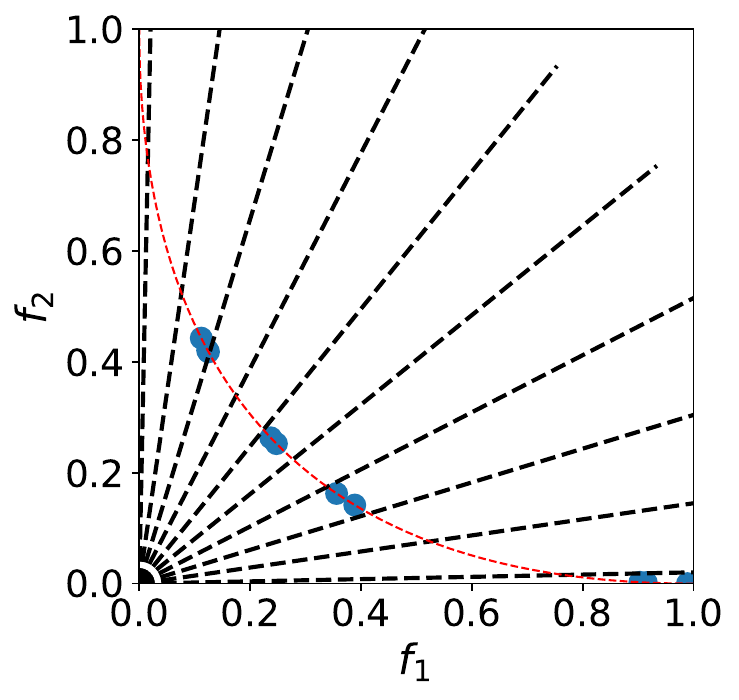}}
\hfill
\subfloat[PMTL \cite{PMTL}.]{\includegraphics[width=0.18\linewidth]{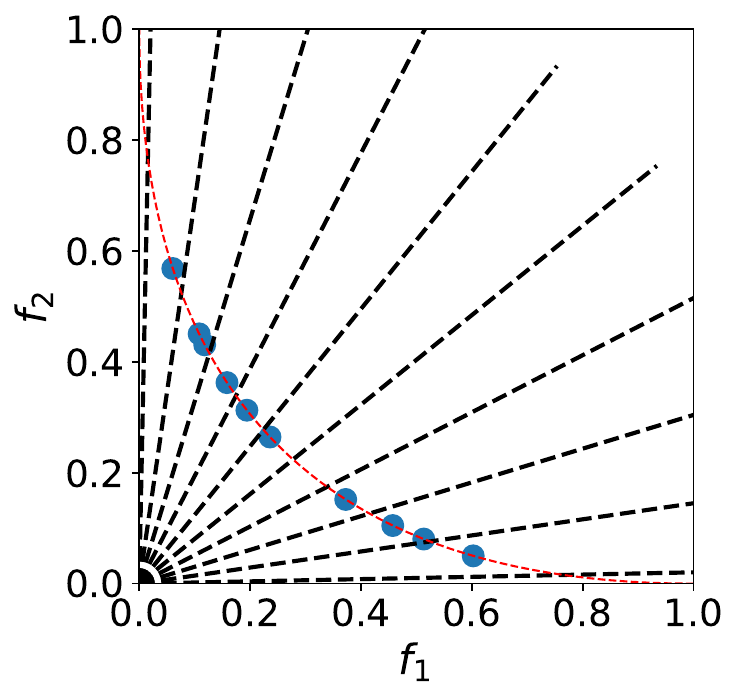}}
\hfill
\subfloat[EPO \cite{EPO, EPO2}.]{\includegraphics[width=0.18\linewidth]{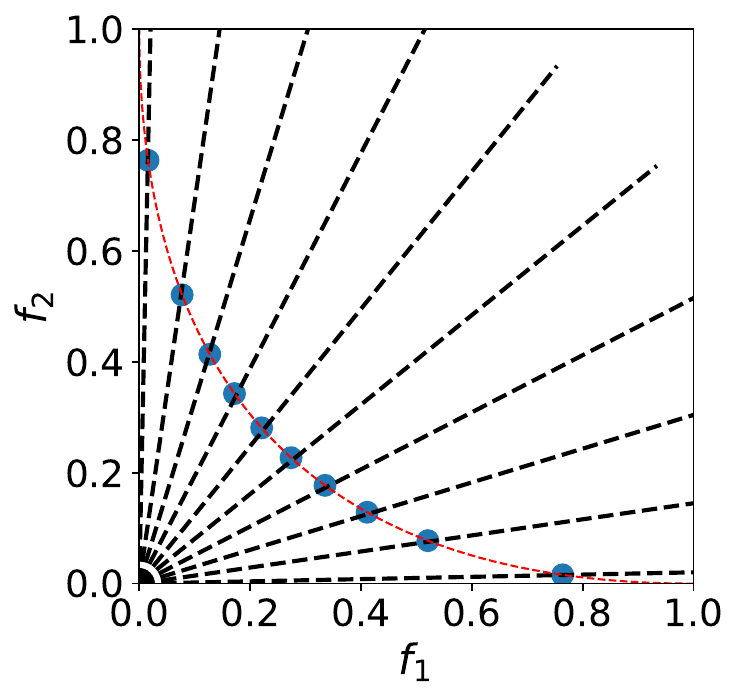}}
\hfill
\subfloat[GradHV \cite{wang2017hypervolume}.]{\includegraphics[width=0.18\linewidth]{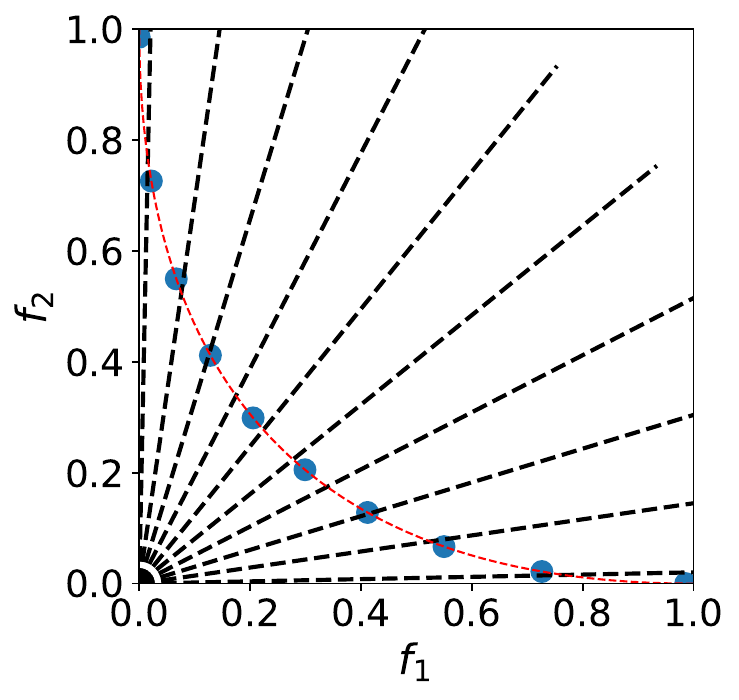}} 
\hfill
\subfloat[MOO-SVGD \cite{liu2021profiling}.]{\includegraphics[width=0.18\linewidth]{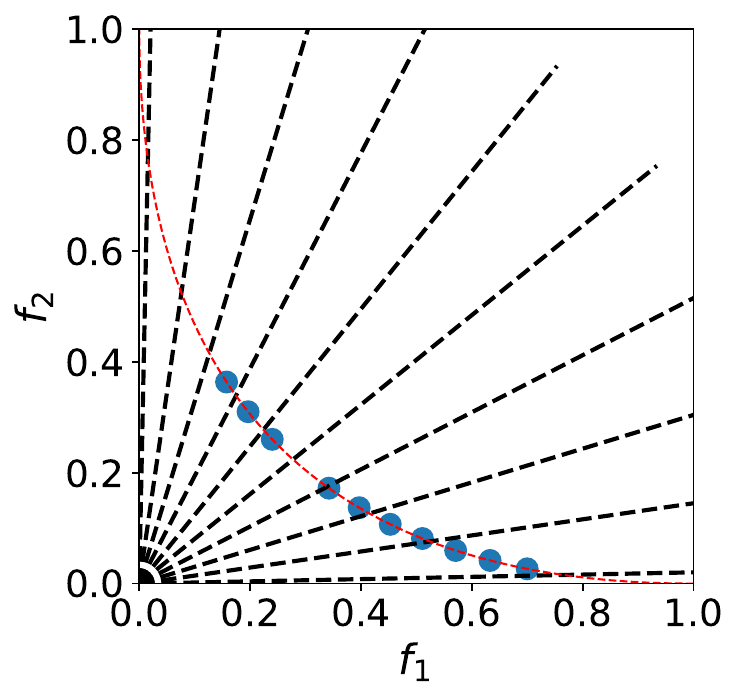}}

\caption{
Distributions of solutions obtained by using representative methods: PMTL~\cite{PMTL} and EPO~\cite{EPO, EPO2} are preference vector-based methods, while GradHV~\cite{wang2017hypervolume} and MOO-SVGD~\cite{liu2021profiling} are methods not using preference vectors. The blue circles denote the solutions, the red curves denote the ground truth PF and the black lines denote the preference vectors.
As can be seen, MGDA~\cite{mtlasmoo} demonstrates poor diversity due to its lack of preference incorporation. While the preference vector-based method PTML~\cite{PMTL} achieves better alignment with the preference vector, it still falls short of exact alignment. In contrast, EPO~\cite{EPO, EPO2} successfully achieves exact alignment. GradHV \cite{wang2017hypervolume} and MOO-SVGD \cite{liu2021profiling} can also generate diverse solutions, despite they do not use a preference vector.
}
\label{fig:mtl_finite} 
\end{figure*}

\section{Finding a Finite Set of Solutions}
\label{sec:pareto_set}
In some scenarios, a single solution may be insufficient to balance the objectives, as users may prefer to obtain multiple trade-off solutions and select one based on their specific needs. This section introduces algorithms to identify a finite set of solutions, providing a discrete approximation of the Pareto set. We first discuss 
methods based on preference vectors (Section~\ref{sec:preference_aware}), followed by 
approaches that do not require the use of preference vectors (Section~\ref{sec:preference_free}). Figure~\ref{fig:mtl_finite} shows solutions obtained by some representative methods on LSMOP1~\cite{cheng2016test} after 5000 iterations of each algorithm.

\subsection{Methods Based on
	Preference Vectors}
\label{sec:preference_aware}
Preference vector-based methods rely on a preference vector set \(\{\valp^{(1)}, \dots, \valp^{(n)}\}\). These preference vectors partition the problem into $n$ subproblems, where each subproblem involves finding the solution that corresponds to a specific preference vector in the set. By solving $n$ subproblems, these methods obtain a set of $n$ solutions.

Pareto Multi-Task Learning (PMTL)~\cite{PMTL}, inspired by the idea of decomposition-based MOO algorithms~\cite{zhang2007moea}, incorporates preference vectors as constraints on the objectives. For the \(i\)-th subproblem with preference vector \(\valp^{(i)}\), the objective vector \(\vf(\vtheta)\) is constrained to be closer to \(\valp^{(i)}\) than to the other preference vectors:
\begin{align}
	\min_{\vtheta \in \gK \subset \R^d}\vf(\vtheta) := [f_1(\vtheta), \dots, f_m(\vtheta)]^\top, ~~\mathrm{s.t.} ~~ r^{(j)}(\vtheta) := (\valp^{(j)} - \valp^{(i)})^\top \vf(\vtheta) \leq 0, ~~ j \in [n].
	\label{eq:mooc}
\end{align}
Let \(\mR\) be the matrix containing gradients of the active constraints,
i.e, \(\mR = \big[\nabla r^{(j)}(\vtheta)\big]_{j \in \sI(\vtheta)}\), with \(\sI(\vtheta) = \{j \in [n] \mid r^{(j)}(\vtheta) \geq -\epsilon\}\) containing
indices
of the active constraints. Using a method akin to MGDA \cite{mtlasmoo}, PMTL derives a descent direction for problem (\ref{eq:mooc}) that optimizes all \(m\) objectives while keeping \(\vf(\vtheta)\) within the sector closer to the preference vector \(\valp^{(i)}\). The descent direction is:
\begin{align}
	\vd = - \mG \vlam + \mR \vbeta,
\end{align}
where $\mG=[\nabla f_1(\vtheta), \dots, \nabla f_m(\vtheta)]$ is the Jacobian matrix. $\vlam$ and $\vbeta$ are coefficients obtained by solving the following quadratic programming problem:
\begin{align}
	\min_{\vlam, \vbeta} \norm{\mG \vlam + \mR \vbeta}^2, ~~\text{s.t.}~~ \vlam^\top\bm{1} + \vbeta^\top\bm{1} = 1, \; \lambda_i \geq 0, \; \beta_j \geq 0.
\end{align}
A limitation of PMTL is it can only constrain objective vectors within sectors, which lacks precise control over the position of Pareto solutions.

Exact Pareto Optimal search (EPO)~\cite{EPO, EPO2} is designed to locate Pareto solutions which are aligned the objective vector exactly with given preference vectors. 
EPO achieves this using a uniformity function, \(u_{\valp}(\vf(\vtheta)) = \text{KL}(\hat{\vf}(\vtheta) \parallel \nicefrac{\bm{1}}{m})\), where:
$
\hat{f}_i(\vtheta) = \frac{\alpha_i f_i(\vtheta)}{\sum_{j=1}^m \alpha_j f_j(\vtheta)}.
$
Minimizing this function aligns \(\vf(\vtheta)\) with \(\valp\), achieving precise alignment and Pareto optimality.
Let \(\mC = \mG^{\top}\mG\) and \(\vc_i\) be the \(i\)-th column of \(\mC\). Let $\va = [a_1, \dots, a_m]^\top$, where $a_i = \alpha_i(\log(m \hat{f}_i(\vtheta)) - u_\valpha(f(\vtheta)))$. At each iteration, EPO classifies objective indices into three sets:
\(\sJ = \{j \mid \va^\top \vc_j > 0\}\) are the indices decreasing uniformity, \(\overline{\sJ} = \{j \mid \va^\top \vc_j \leq 0\}\) are the indices increasing uniformity, and \(\sJ^*\) are the indices with the maximum \(\alpha_j f_j(\vtheta)\). Based on these sets, it determines \(\vlam\) for the common descent direction $\vd = \mG \vlam$ by solving:
\begin{align}
	\max_{\vlam \in \simplex}\vlam^\top \mC (\va \mathbbm{1}_{u_{\valp}} + \bm{1}(1 - \mathbbm{1}_{u_{\valp}})), ~~\mathrm{s.t.} \left \{
	\begin{array}{ll}
		\vlam^\top \vc_j \geq \va^\top \vc_j \mathbbm{1}_\sJ, \quad \forall j \in \overline{\sJ} \setminus \sJ^*, \\
		\vlam^\top \vc_j \geq 0, \quad \forall j \in \sJ^*.
	\end{array}
	\right.
	\label{eqn:lp}
\end{align}
This updating direction can balance objectives while reducing non-uniformity, resulting in solutions that exactly match each preference vector. However, EPO has three drawbacks: (1) objective vectors must be non-negative, (2) unnecessary complexity from dividing sets into three subsets, and (3) computational inefficiency due to repeated Jacobian calculations and solving linear programming problem.

To address these drawbacks, Weighted Chebyshev MGDA (WC-MGDA)~\cite{WCMGDA} considers the dual form of Tchebycheff scalarization (problem (\ref{eq:tch})) of a linear programming problem.
Optimizing Tchebycheff function addresses the previously mentioned drawbacks of EPO. However, its convergence rate is relatively slow. Two methods for improving the Tchebycheff scalarization include smooth Tchebycheff scalarization, which replaces the non-smooth max operation with a smooth approximation, and Preference-based MGDA (PMGDA), which only requires the update direction to have a negative inner product with the exact constraint gradient.

FERERO~\cite{chenferero} captures preference information within the MGDA framework, addressing both objective constraints with constants and the exactness constraint. 
It achieves fast convergence rates of $\mathcal{O}(\epsilon^{-1})$ for deterministic gradients and $\mathcal{O}(\epsilon^{-2})$ for stochastic gradients, where $\epsilon$ is the error tolerance.

The methods discussed above rely on a fixed set of preference vectors. However, in real-world applications where the shape of the Pareto front is unknown, a predefined set of preference vectors may not always result in well-distributed solutions. To overcome this limitation, GMOOAR~\cite{gmooar} formulates the problem as a bi-level optimization problem. In the upper level, preference vectors are optimized to maximize either the hypervolume or uniformity of the solutions. In the lower level, solutions are optimized based on these preference vectors. This approach enables the algorithm to dynamically adjust preference vectors based on the given optimization problem, ensuring the desired solution distribution.
UMOD~\cite{zhang2024gliding} introduces an approach that aims to maximize the minimum distance between solutions: 
\begin{equation}
	\label{eqn:maxmin}
	\max_{\vtheta^{(i)}, \vtheta^{(j)} \in \mathrm{PS}} \min_{1 \leq i < j \leq n} \rho( \vf(\vtheta^{(i)}), \vf(\vtheta^{(j)})),
\end{equation}
where $\rho(\cdot, \cdot)$ denotes the Euclidean distance. UMOD achieves desirable solution distributions with theoretical guarantees: (1) For bi-objective problems with a connected, compact PF, the optimal solution includes the PF endpoints, with equal distances between neighboring objective vectors; (2) As the number of solutions increases, the objective vectors asymptotically distribute on the PF.

\subsection{Methods without Using 
	Preference Vectors}
\label{sec:preference_free}
Unlike methods that rely on preference vectors, another approach directly optimizes for a diverse set of Pareto-optimal solutions. As HV evaluates solution sets based on both convergence and diversity, it is commonly used in traditional evolutionary MOO~\cite{shang2020survey}. In the context of gradient-based MOO, gradient-based hypervolume maximization algorithms (GradHV)~\cite{wang2017hypervolume,deist2021multi,deist2020moo,emmerich2007gradient} have been developed. Let
$\{\vtheta^{(1)}, \dots, \vtheta^{(n)}\}$ 
be a set of 
solutions and 
$\{\vf(\vtheta^{(1)}), \dots, \vf(\vtheta^{(n)})\}$ 
be their corresponding objective vectors. These algorithm first calculate the hypervolume gradient of the $i$-th solution $\vtheta^{(i)}$ as follows:
\begin{equation}
	\label{eq:hv}
	\vd^{(i)} = \sum_{k=1}^m \underbrace{\frac{\partial \mathrm{HV}_\vr(\{ \vf(\vtheta^{(1)}), \dots, \vf(\vtheta^{(n)})\})}{\partial f_k(\vtheta^{(i)})}}_{a_i : 1 \times 1} \underbrace{\frac{\partial f_k(\vtheta^{(i)})}{\partial \vtheta^{(i)}}}_{\vB: 1 \times d}.
\end{equation}
Once $\vd^{(i)}$ is calculated, solutions $\vtheta^{(i)}$'s are updated by gradient ascent, $\vtheta^{(i)} \leftarrow \vtheta^{(i)}+\eta \vd^{(i)}$, where $\eta$ is a learning rate. 
The term $\vB$ in Equation (\ref{eq:hv}) can be easily estimated via backpropagation. The main challenge lies in estimating the first term $a_i$. As proposed by Emmerich et al.~\cite{emmerich2007gradient}, index sets are categorized based on differentiability: $\sZ$ includes subvectors with zero partial gradients, \(\sU\) contains subvectors with undefined or indeterminate gradients, and \(\sP\) has subvectors with positive gradients. This classification leads to many zero-gradient terms, improving efficiency. Then, non-zero terms are computed using a fast dimension-sweeping algorithm~\cite{beume2009complexity, guerreiro2012fast}. This method applies to bi-, tri-, and four-objective problems with time complexities of \(\Theta(mn + n \log n)\) for $m=2$ or $3$ and \(\Theta(mn + n^2)\) for $m=4$, where \(m\) is the number of objectives and \(n\) is the number of solutions. To further enhance convergence, the Newton-hypervolume method~\cite{sosa2020set} incorporates second-order information for faster updates. Deist et al. ~\cite{deist2021multi} apply the hypervolume gradient methods for deep learning tasks such as medical image classification with more than thousands of parameters.

Inspired by Stein Variational Gradient Descent (SVGD) \cite{liu2016stein},
MOO-SVGD \cite{liu2021profiling} proposes a different approach 
to maintain diversity while enhancing convergence. The update direction for the $i$-th solution $\vtheta^{(i)}$ is given by:
\begin{align}
	\vd^{(i)} = & - \sum_{j=1}^n \vg^*(\vtheta^{(j)}) \kappa(\vf(\vtheta^{(i)}), \vf(\vtheta^{(j)}))\nonumber + \mu \nabla_{\vtheta^{(i)}} \kappa(\vf(\vtheta^{(i)}), \vf(\vtheta^{(j)})). \label{eq:svgd}
\end{align}
where $\vg^*(\vtheta^{(j)}) = \sum_{i=1}^m \lambda_i^* \nabla f_i(\vtheta^{(j)}),
$ and \(\{\lambda_i^*\}_{i=1}^m\) are the objective weights obtained by MGDA \cite{mtlasmoo} (i.e., problem (\ref{eq:mgda})). 
\(\kappa(\cdot, \cdot)\) represents a kernel function, often chosen as the radial basis function (RBF) kernel. The first term in Equation~(\ref{eq:svgd}) drives the particles toward the target distribution, while the second term acts as a repulsive force that promotes diversity.

While the methods discussed above address scenarios with more solutions than objectives ($n>m$), recent work has begun to explore the converse case, where objectives outnumber solutions ($m>n$) \cite{lin2024few,li2024many,liu2024many,ding2024efficient}.
These approaches assign each objective to a solution $\vtheta \in \{\vtheta^{1}, \dots, \vtheta^{n}\}$ that achieves the smallest objective value for that particular objective.
The goal is then to minimize an aggregation of these smallest objective values, typically either their sum or their maximum. The Few for Many (F4M) framework~\cite{lin2024few,liu2024many}, for example, employs a minimax strategy by minimizing the maximum value, which is formulated as:
\begin{align}
	\min_{\vtheta^{1}, \dots, \vtheta^{n}} \max_{i \in [m]}  \min_{\vtheta \in \{\vtheta^{1}, \dots, \vtheta^{n}\}} f_i(\vtheta).
\end{align}

Different from F4M, Sum of Minimum (SoM) \cite{ding2024efficient} uses the sum aggregation, formulated as: 
\begin{align}
	\min_{\vtheta^{1}, \dots, \vtheta^{n}} \sum_{i=1}^m \left ( \min_{\vtheta \in \{\vtheta^{1}, \dots, \vtheta^{n}\}} f_i(\vtheta) \right ).
\end{align}

{MosT}~\cite{li2024many} frames the problem as a bilevel optimization task, with the outer loop using MGDA to optimize solutions and the inner loop solving an optimal transport problem to assign objectives to solutions.

\subsection{Discussions}
This section discusses two primary approaches for obtaining multiple Pareto optimal solutions. Preference vector-based methods decompose a MOP into multiple independent subproblems based on predefined preference vectors. These subproblems can be solved independently, which enhances memory efficiency and enables parallel computation. However, in real-world applications, determining suitable preference vectors can be challenging due to the complex and often unpredictable nature of the Pareto front. Furthermore, since the subproblems are solved independently, they cannot benefit from shared information between one another.

On the other hand, methods not using preference vectors bypass the challenge of selecting preference vectors by directly optimizing a set of solutions. These methods generate a set of Pareto-optimal solutions that inherently share interdependent information. However, this approach typically requires higher memory resources and incurs greater computational costs, making it less efficient in resource-constrained scenarios.

\section{Finding an Infinite Set of Solutions}
\label{sec:pareto_infinite}

While a finite set of solutions can only provide a discrete approximation of the Pareto front, many applications require the ability to obtain solutions corresponding to any user preference, effectively representing the entire Pareto set. Directly learning an infinite number of solutions individually is impractical. 
Ma et al. \cite{ma2020efficient} initially investigated deriving an infinite set of solutions by approximating the Pareto set
with first-order expansion around discrete Pareto-optimal solutions.
However, this method has several drawbacks: (1) The approximation error grows when the solutions are widely spaced, (2) First-order approximations perform poorly in high-dimensional objective spaces, and (3) The approach requires solving a linear programming problem for updates, which can be computationally challenging.
To address these issues, many methods have been introduced that leverage neural networks to learn mappings from user preferences to solutions directly, enabling the capture of the entire PS. These methods rely on designing efficient network architectures and implementing effective training strategies.

\begin{figure*}[!t]
\centering
\begin{subfigure}{0.3\linewidth}
\centering
\includegraphics[width=\textwidth]{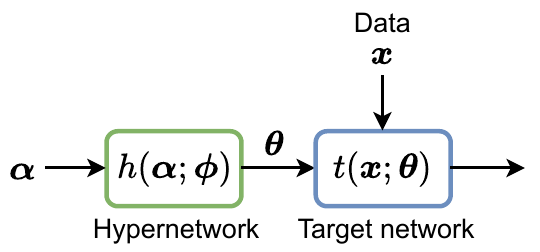}
\caption{Methods based on the hypernetwork.}
\label{fig:hypernetwork}
\end{subfigure}
\hfill
\begin{subfigure}[b]{0.37\linewidth}
\centering
\includegraphics[width=\textwidth]{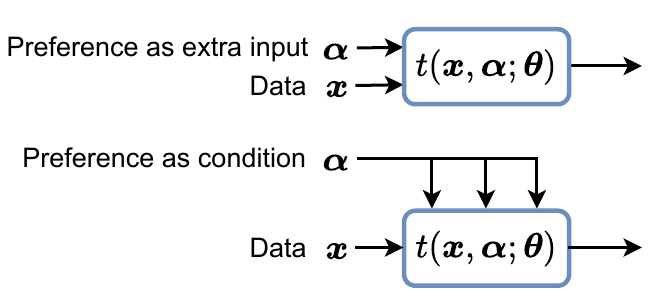}
\caption{Methods based on preference-conditioned network.}
\label{fig:condition1}
\end{subfigure}
\hfill
\begin{subfigure}[b]{0.28\linewidth}
\centering
\includegraphics[width=\textwidth]{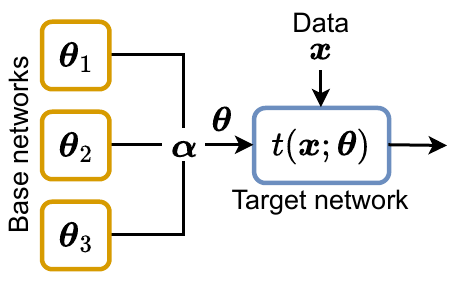}
\caption{Methods based on model combination.}
\label{fig:combination}
\end{subfigure}
\caption{Illustration of different structures to learn an infinite number of solutions.}
\label{fig:struct_illus}
\end{figure*}

In Section \ref{sec:network_structures}, we introduce various network structures designed to capture the entire PS through preference-based mappings. Specifically, current algorithms mainly use three types of network structures: (1) Hypernetwork in Section \ref{sec:hypernetwork}, (2) Preference-Conditioned Network in Section \ref{sec:pref_net}, and (3) Model Combination in Section \ref{sec:combine}. An illustration of these methods is shown in Figure \ref{fig:struct_illus}. Section \ref{sec:training_strategy} discusses training strategies for these network structures.

\subsection{Network Structure}
\label{sec:network_structures}

\subsubsection{Hypernetwork}
\label{sec:hypernetwork}
Hypernetwork is a neural network that generates the parameters for another target network \cite{ha2016hypernetworks}. This idea has been leveraged by the Pareto Hypernetwork (PHN) \cite{PHN} and Controllable Pareto Multi-Task Learning (CPMTL) \cite{cpmtl} to learn the entire Pareto set, where the input is the user preference $\valp$ and the output is the target network's parameters. The hypernetwork usually consists of several MLP layers, and has been adopted in many subsequent approaches such as PHN-HVI \cite{PHNHVI} and SUHNPF \cite{SUHNPF}. Recently, Tuan et al. \cite{tuan2024hyper} propose using a transformer architecture as the hypernetwork, demonstrating superior performance compared to MLP-based hypernetwork models.

A primary limitation of hypernetwork-based algorithms is their sizes. Since the output dimension matches the number of parameters in the target network, hypernetworks are often much larger than the base networks, limiting their applicability to large models. Some methods address this limitation by employing hypernetworks with chunking \cite{PHN, cpmtl}. Chunking involves partitioning the parameter space into smaller, more manageable segments, enabling the hypernetwork to generate parameters more efficiently and scalability. This approach reduces the overall size of the hypernetwork while preserving its capability to produce accurate parameters for the target network.

\subsubsection{Preference-Conditioned Network}
\label{sec:pref_net} 
Instead of using a hypernetwork to generate weights, the original model can be directly modified to incorporate preferences. COSMOS~\cite{COSMOS} suggests adding preferences as an additional input to the model by combining user preference $\valp$ with input data $\vx$, thereby increasing the input dimension of the original model.

However, such input-based conditioning has limited ability to produce diverse solutions.
Some studies~\cite{dosovitskiy2019you, gmooar, raychaudhuri2022controllable} employ the Feature-wise Linear Modulation (FiLM) layer~\cite{perez2018film} to condition the network. The FiLM layer works by applying an affine transformation to the feature maps. Specifically, given a feature map $\vu$ with $C$ channels, we use an MLP to generate conditioning parameters $\vgamma \in \R^C$ and $\vbeta \in \R^C$ based on the given preference $\valp$. Then, the FiLM layer modifies the features as $
\vu'_c = \gamma_c \cdot \vu_c + \beta_c,$
where $\vu_c$ is the $c$-th channel of $\vu$. This method achieves better conditioning ability compared to merely adding the preference to the input layer. Raychaudhuri et al. \cite{raychaudhuri2022controllable} also propose leveraging another network to predict the network architecture. It dynamically adapts the model's structure according to user preferences, allowing for more flexible and efficient learning.

\subsubsection{Model Combination}
\label{sec:combine}
Methods 
based on model combination
construct a composite model by integrating multiple individual models, thereby offering an effective way to introduce diversity. PaMaL \cite{PAMAL} achieves this by learning several base models and combining them through a weighted sum of their parameters based on user preferences:
\begin{equation}
	\vtheta(\valp) = \sum_{i=1}^m \alpha_i \vtheta_i.
\end{equation}
Although this incorporates more parameters than 
methods based on preference-conditioned networks, PaMaL enhances diversity and provides users with a broader range of choices.

Despite its benefits, learning multiple models simultaneously may become inefficient when handling a large number of objectives. To mitigate this issue, LORPMAN~\cite{LORPMAN} proposes learning a full-rank model $\vtheta_0\in\mathbb{R}^{p\times q}$ and $m$ low-rank models $\mB_i \mA_i$'s, where $\mB_i \in \mathbb{R}^{p \times r}$ and $\mA_i \in \mathbb{R}^{r \times q}$. Given user preference $\valp$, the composite model is given by:
\begin{equation}
	\vtheta(\valp) = \vtheta_0 + s \sum_{i=1}^m \alpha_i \mB_i\mA_i,
\end{equation}
where $s$ is a scaling factor that adjusts the impact of the low-rank components. This approach efficiently scales Pareto set learning to manage up to $40$ objectives, achieving superior performance. A similar approach is also mentioned in \cite{dimitriadis2024pareto}. To further enhance parameter efficiency, Chen and Kwok \cite{chen2024you} introduce a preference-aware tensor multiplication:
\begin{align}
	\label{eq:td}
	\vtheta(\valp) = \vtheta_0 + s \mC \times_1 \mA \times_2 \mB \times_3 \valp,
\end{align}
where $\mC \in \mathbb{R}^{r \times r \times m}$ is a learnable core tensor, $\mA \in \mathbb{R}^{r \times p}$, $\mB \in \mathbb{R}^{r \times q}$ are learnable matrices, and $\times_u$ denotes mode-$u$ tensor multiplication~\cite{kolda2009tensor}.
Zhong et al.~\cite{panacea} address scalability challenges for large-scale models, such as large language models, using SVD-LoRA:
\begin{equation}
	\vtheta(\valp) = \vtheta_0 + \mU \mSigma \mV,
\end{equation}
where $\mSigma$ is a diagonal matrix defined as $\text{diag}(\sigma_1, \dots, \sigma_r, s\alpha_1, \dots, s\alpha_m)$, $\{\sigma_i\}_{i=1}^r$ and $s$ are learnable scalars. $\mU \in \mathbb{R}^{p \times (r+m)}$ and $\mV \in \mathbb{R}^{(r+m) \times q}$ are learnable matrices. 

Instead of training new models, Tang et al.~\cite{tang2024towards} propose to utilize a mixture of experts~\cite{moesurvey} to combine pre-existing models in order to achieve the Pareto set. Given $n$ models, $\vtheta_1, \dots, \vtheta_n$, each trained on different datasets, 
they are first integrated to a unified base model $\vtheta_0$. For some modules in the network, differences between the individual models and the base model (i.e., $\vtheta_i - \vtheta_0$'s) are maintained. These differences are treated as experts within the framework. Then, a
mixture of experts is applied, where a gating network conditioned on user preferences determines the weighting of these experts. It enables the generation of diverse models by applying different weighted combinations of experts to align with specific user preferences.

\subsection{Training Strategy}
\label{sec:training_strategy}
This section outlines the training approach for the parameters 
(denoted $\vphi$)
of the network structures introduced in Section~\ref{sec:network_structures}.
Specifically, $\vphi$ refers to the parameters of the hypernetwork in Section~\ref{sec:hypernetwork}, parameters of the preference-conditioned network in Section~\ref{sec:pref_net}, or parameters of the base models or low-rank models in Section~\ref{sec:combine}. Current algorithms follow a similar approach: sample user preferences and optimize the structure parameters to generate networks that align with those preferences.
The training objective can be written as:
\begin{equation}
	\min_\vphi \E_{\valp \sim \simplex}~\E_{(\vx,\vy) \sim \mathcal{D}} \left[ \tilde{g}_\valp(\vell(t(\vx; \vtheta(\valp; \vphi), \vy)) \right], 
	\label{eqn:psl}
\end{equation}
where $(\vx,\vy)$'s are training data sampled from dataset $\mathcal{D}$, $\vell(\cdot,\cdot)$ is the multi-objective loss function and $\tilde{g}_\valp(\cdot)$ is an MOO algorithm that produces a single solution given preference vector $\valp$. In general, most optimization algorithms discussed in Sections~\ref{sec:scalarization} and \ref{sec:pareto_set} can be adopted. Below, we summarize the algorithms adopted by the existing methods.
\begin{itemize}
	\item Scalarization: As outlined in Section \ref{sec:scalarization}, scalarization combines multiple objectives into one scalar objective. Linear scalarization is common in most algorithms, such as PHN~\cite{PHN}, COSMOS~\cite{COSMOS}, PAMAL~\cite{PAMAL}, and LORPMAN~\cite{LORPMAN}. Alternatives like Tchebycheff and smooth Tchebycheff scalarization~\cite{smooth} can also be adopted.
	
	\item Preference-Aware Weighting Methods: As discussed in Section \ref{sec:preference_aware}, these methods incorporate user preferences into optimization. PHN~\cite{PHN} employs the Exact Pareto Optimal (EPO) solver~\cite{EPO} to find Pareto-optimal solutions aligned with preferences. CPMTL~\cite{cpmtl} uses a constrained approach inspired by PMTL~\cite{PMTL}.
	
	\item Hypervolume Maximization: As discussed in Section~\ref{sec:preference_free}, hypervolume maximization can optimize both diversity and convergence of a solution set. PHN-HVI~\cite{PHNHVI} leverages hypervolume maximization to optimize the structural parameters $\vphi$. In each iteration, it samples a set of preference vectors and generates solutions based on these vectors. Subsequently, hypervolume maximization is applied to promote both the diversity and convergence of the solutions.
	
\end{itemize}

\subsection{Discussions}
Selecting the appropriate structural framework and optimization algorithms is important for effectively learning an infinite Pareto set. Hypernetwork-based approaches offer significant flexibility by generating a diverse set of models tailored to varying user preferences. However, their main drawback is size: the output dimension matches the target network's parameters, making hypernetworks significantly larger than base models. This limits their applicability for large-scale networks or resource-constrained scenarios.

In contrast, methods based on preference-conditioned networks
are highly parameter-efficient. They introduce a minimal number of additional parameters relative to the base model, making them well-suited for large-scale applications where adding extra parameters is costly. However, their limited parameter space may reduce diversity, hindering their ability to approximate complex PS accurately.

Methods based on model combinations serve as a middle ground between hypernetwork and preference-conditioned network methods. By integrating multiple models, they offer a good approximation of the PS while maintaining scalability for large models. Furthermore, some methods allow for parameter adjustments through the selection of appropriate ranks in low-rank approximations.

Regarding training strategies, most algorithms use linear scalarization for its simplicity and strong performance. However, when aligning solutions with user preferences is important, preference-dependent weighting schemes in Section~\ref{sec:preference_aware}
are more suitable. Additionally, to achieve HV-optimal solutions, HV-based optimization criteria in Section~\ref{sec:preference_free} 
can also be employed, as they better capture the PF by focusing on the dominated objective space volume.

\section{Theories}
\label{sec:theory}
\subsection{Convergence of Gradient-Balancing Methods in Section \ref{sec:grad_balancing}} \label{sec:convergence}
This section reviews the convergence analysis for gradient-balancing methods in Section \ref{sec:grad_balancing}. This is first examined under the deterministic gradient setting (i.e., full-batch gradient). Subsequently, it is analyzed under the stochastic gradient setting.
Note that in non-convex MOO, convergence refers to reaching Pareto stationary (Definition \ref{def:stationary}).

\subsubsection{Deterministic Gradient}
With the use of deterministic gradients, it has shown that under mild conditions, MGDA \cite{mtlasmoo} can converge to a Pareto-stationary point at a convergence rate of $\mathcal{O}(K^{-1/2})$~\cite{fliege2019complexity}, where $K$ is the number of iterations. This rate is comparable to that of single-objective optimization~\cite{nesterov2013introductory}. The fundamental idea is to assess the reduction in each individual objective function. When the current solution is not Pareto-stationary, it can be shown that the common descent direction $\vd$ identified by MGDA ensures an improvement of at least $\mathcal{O}(\norm{\vd}^2)$ in each objective, which mirrors the situation in single-objective optimization~\cite{fliege2019complexity}. Therefore, the method used in single-objective optimization can be applied to prove the convergence. Furthermore, other algorithms such as CAGrad~\cite{CAGrad}, Nash-MTL~\cite{NashMTL}, and PCGrad~\cite{pcgrad} also offer convergence analyses in similar approaches.

\subsubsection{Stochastic Gradient}
Liu and Vicente~\cite{SMGDA} are the first to analyze the convergence of gradient-based MOO algorithms with stochastic gradients. They propose a stochastic version of MGDA \cite{mtlasmoo} that computes a common descent direction using stochastic gradients for all objectives and prove convergence to Pareto-optimal solutions
under the assumption of convex objective functions. However, due to the inherent bias of common descent direction $\vd$ 
introduced by stochastic gradient estimations, their analysis requires the use of an increasing batch size that grows linearly with the number of iterations to ensure convergence. This requirement can be impractical in real-world applications, as large batch sizes lead to increased computational costs and memory usage. 

\begin{table}[!t]
\centering
\caption{Convergence of existing stochastic gradient MOO algorithms. ``LS" and ``GS" denote the objectives are $L$-smooth and generalized $L$-smooth, respectively. ``BG" and ``BF" represent the bounded gradient and bounded function value assumptions, respectively. ``Complexity" denotes the sample complexity
achieving $\epsilon$-accurate Pareto stationary point. ``Bounded CA" denotes the bounded conflict-avoidant distance.} 
\vspace{-0.1in}
\resizebox{0.6\linewidth}{!}{
\begin{tabular}{c|cccc}
\toprule
Method & Batch Size & Assumption & Complexity & Bounded CA \\
\midrule
SMG \cite{SMGDA} & $\gO(\epsilon^{-2})$ & LS, BG  & $\gO(\epsilon^{-4})$ & \XSolidBrush \\
CR-MOGM \cite{zhou2022convergence} & $\gO(1)$ & LS, BF, BG & $\gO(\epsilon^{-2})$ & \XSolidBrush \\
MoCo \cite{fernando2023mitigating} & $\gO(1)$ & LS, BF, BG & $\gO(\epsilon^{-2})$ & \XSolidBrush \\
PSMGD \cite{xu2024psmgd}& $\gO(1)$ & LS, BF, BG & $\gO(\epsilon^{-2})$ & \XSolidBrush \\
MoDo \cite{chen2024three} & $\gO(1)$ & LS, BG & $\gO(\epsilon^{-2})$ & \Checkmark \\
SDMGrad \cite{xiao2023direction} & $\gO(1)$ & LS, BG & $\gO(\epsilon^{-2})$ & \Checkmark \\
SGSMGrad \cite{zhang2024convergence} & $\gO(1)$ & GS & $\gO(\epsilon^{-2})$ & \Checkmark\\
\bottomrule
\end{tabular}}
\label{tab:convergence}
\end{table}

To overcome this limitation,
Zhou et al.~\cite{zhou2022convergence} propose Correlation-Reduced Stochastic Multi-Objective Gradient Manipulation (CR-MOGM). It addresses the bias in the common descent direction by introducing a smoothing technique on weight coefficient $\vlam$. In CR-MOGM, $\vlam^{(k)}$ at iteration $k$ is updated using a moving average:
\begin{equation}
	\vlam^{(k)} = (1 - \gamma) \hat{\vlam}^{(k)} + \gamma \vlam^{(k-1)},
\end{equation}
where $\gamma \in [0, 1]$ is a smoothing factor, $\hat{\vlam}^{(k)}$ is the weight vector obtained by the MOO solver (such as MGDA) at iteration $k$ using the current stochastic gradients, and $\vlam^{(k-1)}$ is the smoothed weight from the previous iteration. Smoothing reduces the variance in the weight, leading to a more stable and reliable common descent direction. Xu et al. \cite{xu2024psmgd} further analyze the convergence by updating the objective weights every $\tau$ iterations instead of at each iteration, and achieve the same convergence rate $\gO(\epsilon^{-2})$ as CR-MOGM. They also match the convergence rate in stochastic single-objective optimization~\cite{ghadimi2013stochastic}. 

Another approach to address gradient bias is MoCo \cite{fernando2023mitigating}, which introduces a tracking variable $\hat{\vg}_{i}^{(k)}$ that approximates the true gradient:
\begin{equation}
	\hat{\vg}^{(k+1)}_i = \prod_{L_i} \left(\hat{\vg}^{(k)}_i - \gamma \left(\hat{\vg}^{(k)}_i - \vg^{(k)}_i\right)\right),
\end{equation}
where $\prod_{L_i}$ is the projection to the set $\{\vg \in \R^d~~|~~\norm{\vg} \leq L_i\}$, and $L_i$ is the Lipschitz constant of $f_i(\vtheta)$.
However, the analysis requires the number of iterations $K$ to be sufficiently large (in the order of $\gO(m^{10})$). This high dependency on $m$ makes MoCo less practical for problems involving many objectives. Additionally, the convergence analysis of CR-MOGM and MoCo relies on the assumption that the objective functions have bounded values.

To address these limitations, Chen et al.~\cite{chen2024three} propose the Multi-objective gradient with double sampling algorithm (MoDo). MoDo mitigates bias in stochastic gradient-based MOO methods without assuming bounded function values. It introduces a double sampling technique to obtain unbiased estimates of the gradient products needed for updating the weight coefficients. Specifically, at each iteration, MoDo updates the weight vector $\vlam^{(k)}$ using gradients computed on two independent mini-batches:
\begin{equation}
	\vlam^{(k+1)} = \prod_{\simplex} \left( \vlam^{(k)} - \eta \mG^{(k)}({\vz_1^{(k)}})^\top \mG^{(k)}({\vz^{(k)}_2}) \vlam^{(k)} \right),
\end{equation}
where $\eta$ is the step size, $\prod_{\simplex}$ is the projection onto the simplex $\simplex$, and $\mG^{(k)}({\vz^{(k)}_1})$, $\mG^{(k)}({\vz^{(k)}_2})$ are stochastic gradients evaluated on two independent mini-batches $\vz_{1}^{(k)}$ and $\vz_{2}^{(k)}$. By reducing the bias in estimating the common descent direction, MoDo achieves convergence guarantees without the bounded function value assumption. Additionally, it guarantees a bounded conflict-avoidant (CA) distance, which is the distance between the estimated update direction and the CA direction defined in problem (\ref{eq:mgda_org}).

Similarly, Xiao et al.~\cite{xiao2023direction} propose the Stochastic Direction-oriented Multi-objective Gradient descent (SDMGrad) algorithm, which introduces a new direction-oriented multi-objective formulation by regularizing the common descent direction within a neighborhood of a target direction (such as the average gradient of all objectives). In SDMGrad, the weight vector $\vlam^{(k)}$ is updated as:
\begin{align}
	& \vlam^{(k+1)} = \prod_{\simplex} \left( \vlam^{(k)} - \eta \left[ \mG^{(k)}({\vz^{(k)}_1})^\top \left( \mG^{(k)}(\vz_2^{(k)}) \vlam^{(k)} + \gamma \vg_0(\vz^{(k)}_2) \right) \right] \right),
\end{align}
where $\eta$ is the step size, $\gamma$ is a regularization factor controlling the proximity to the target direction $ \vg_0(\vz_{2}^{(k)})$. By incorporating this regularization, SDMGrad effectively balances optimization across objectives while guiding the overall descent direction. 
Zhang et al.~\cite{zhang2024convergence} further consider the convergence under generalized $L$-smoothness \cite{li2024convex} and without bounded gradient assumption. A summary of the convergence results of existing stochastic gradient MOO algorithms is provided in Table~\ref{tab:convergence}. 

\vspace{-0.1in}
\subsection{Generalization} \label{sec:theroy_generalization}
The generalization aspect of multi-objective deep learning remains relatively underexplored compared to its convergence properties. Cortes et al. \cite{cortes2020agnostic} analyze the generalization behavior of a specific scalarization approach that minimizes over convex combinations. Sukenik et al. \cite{sukenik2024generalization} consider 
a broader class of scalarization methods. They show that the generalization bounds for individual objectives extend to MOO with scalarization. Both studies rely on the Rademacher complexity of the hypothesis class to establish algorithm-independent generalization bounds, which are unaffected by the training process. 

Another line of research investigates Tchebycheff scalarization, which focuses on the sample complexity required to achieve a generalization error within $\epsilon$ of the optimal objective value. Formally, given a set of $m$ distributions $\{\gD_i\}_{i=1}^m$ and a hypothesis class $\gH$, the goal is to find a (possibly randomized) hypothesis $h$ such that 
\begin{equation}
	\max_{i\in[m]}\ell_{\gD_i}(h) \leq \min_{h^*\in\gH}\max_{i\in[m]}\ell_{\gD_i}(h^*) + \epsilon,
\end{equation}
where $\ell_{\gD_i}(h)$ is the loss 
of hypothesis $h$
on $\gD_i$,
$h^*$ is the optimal hypothesis and $\epsilon$ is the error tolerance. It is first formulated by Awasthi et al.~\cite{awasthi2023open} as an open problem in 2023 and has since been addressed by several works~\cite{peng2024sample, zhang2024optimal}. 
Haghtalab et al. are the first to show the 
sample complexity 
lower bound 
of $\widetilde{\Omega}\left(\frac{v + m}{\epsilon^2}\right)$, where $v = \VCdim(\gH)$ is the VC-dimension of the hypothesis class $\gH$. By using a boosting framework, Peng~\cite{peng2024sample} gives an algorithm that achieves a sample complexity upper bound 
of $\widetilde{O}\left(\frac{v + m}{\epsilon^2}\cdot \left(\frac{m}{\epsilon}\right)^{o(1)}\right)$, which is nearly optimal. In a concurrent work, Zhang et al.~\cite{zhang2024optimal} provide a variant of the hedge algorithm under an Empirical Risk Minimization (ERM) oracle access that achieves the optimal sample complexity bound of $\widetilde{O}\left(\frac{v + m}{\epsilon^2}\right)$.

In the online setting, Liu et al.~\cite{liu2024online} provide an adaptive online mirror descent algorithm that achieves 
$\mathcal{O}\left(\frac{mv}{\sqrt{K}}\right)$  regret,
where $K$ is the number of iterations. Using a plain online-to-batch conversion scheme, this algorithm leads to an $\mathcal{O}\left(\frac{m^2v^2}{\epsilon^2}\right)$ sample complexity, which still lags behind the optimal offline sample complexity of $\widetilde{O}\left(\frac{m+v}{\epsilon^2}\right)$~\cite{zhang2024optimal}. It is interesting to see whether it is possible to use an online learner with proper online-to-batch conversion schemes that is able to match the optimal offline sample complexity. 

MoDo \cite{chen2024three} introduces a different perspective by leveraging algorithm stability to derive algorithm-dependent generalization error bounds of gradient balancing algorithms.

\vspace{-0.1in}
\section{Applications}
\label{sec:application}
\subsection{Reinforcement Learning}
\label{sec:rl}
Reinforcement Learning (RL)~\cite{wang2024rl} involves training agents to make sequential decisions by maximizing cumulative rewards. Multi-Objective Reinforcement Learning (MORL)~\cite{felten2024multi} extends single-objective RL by utilizing a vector-valued reward function \(\vr(s,a): \mathcal{S} \times \mathcal{A} \mapsto \mathbb{R}^m\). MORL aims to learn a policy network \(\pi_\vtheta(s)\) with the following objective:
\begin{equation}
	\label{eqn:morl:expression}
	\min_{\vtheta}\vf(\vtheta) := \mathbb{E}_{a \sim \pi_\vtheta(s)} \sum_{t=1}^{\infty} \gamma^t \vr(s_t, a_t),
\end{equation}
where $t$ is the time step and \(\gamma^t\) is the discount factor.

Radius Algorithm (RA)~\cite{parisi2014policy} employs linear scalarization with fixed preference vectors to transform a MORL problem into a single-objective RL problem. However, a uniform distribution of preferences does not always lead to a uniform distribution of solutions. To address this, various methods focus on improving the configuration of preference vectors. For instance, Optimal Linear Support (OLS)~\cite{roijers2016multi, roijers2014linear} adds new preference vectors dynamically during each iteration. Prediction-guided MORL (PGMORL)~\cite{xu2020prediction} employs a hyperbolic model to map preference vectors to objective values, periodically updating preference vectors to optimize both the hypervolume and a diversity indicator.

Although Lu et al.~\cite{lu2023multi} demonstrated that linear scalarization can recover the entire PS when the policy space includes all stochastic policies, this assumption may be too strong in practice, necessitating approaches beyond naive linear scalarization. CAPQL~\cite{hu2023revisiting} introduces a scalarization method that combines linear scalarization with an entropy term. PPA~\cite{kyriakis2022pareto} incorporates a term to align objective values more closely with preference vectors. Several studies~\cite{peng2023nonlinear, van2013scalarized, bai2022joint, van2014multi, van2013hypervolume, wang2024multi} have explored other scalarization functions, such as the Tchebycheff function, to identify Pareto-optimal policies. Beyond scalarization, gradient balancing methods (discussed in Section~\ref{sec:grad_balancing}) also demonstrate promising performance in MORL.

Another research direction focuses on learning the entire Pareto set~\cite{pirotta2015multi, lin2022pareto_combinatorial, basaklar2022pd, zhu2023scaling, shu2024learning, liu2025pareto}. In these approaches, the policy network \(\pi_\vtheta(s, \valp)\) is conditioned on both the state \(s\) and preference vector \(\valp\). Most of these methods employ training strategies similar to those described in Section~\ref{sec:training_strategy}, but with an RL-specific training objective.

\subsection{Bayesian Optimization}
\label{sec:bo}
Bayesian Optimization (BO)~\cite{bo_review} is a framework for optimizing expensive-to-evaluate black-box functions by using a probabilistic surrogate model. 
When extended to handle multiple black-box objectives simultaneously, this approach is known as Multi-Objective Bayesian Optimization (MOBO)~\cite{knowles2006parego}. 
In MOBO, acquisition functions are designed to manage the balance between exploration and exploitation across \textit{multiple} objectives. Based on this, MOBO methods can be classified into three main categories: (i) \textbf{Scalarization-based methods} convert MOO problems into single-objective ones using scalarization functions discussed in Section \ref{sec:scalarization}~\cite{knowles2006parego, zhang2009expensive}. Zhang and Golovin~\cite{zhang2020random} propose using the hypervolume indicator for scalarization. Instead of scalarizing first and then applying single-objective acquisition functions, methods like SMS-EGO~\cite{ponweiser2008multiobjective}, PAL~\cite{zuluaga2013active, zuluaga2016pal}, and MOBO-RS~\cite{paria2020flexible} extend single-objective acquisitions to multi-objective optimization. (ii) \textbf{Hypervolume improvement-dased methods} extend Expected Improvement (EI) ~\cite{jones1998efficient} to Expected Hypervolume Improvement (EHVI)~\cite{emmerich2006single}, which selects candidate points to enhance the current solution set's hypervolume. Many algorithms are proposed to improve the efficiency and allow parallel candidate generation ~\cite{yang2019efficient, daulton2020differentiable, yang2019multi, konakovic2020diversity, daulton2021parallel, daulton2022multi, daulton2023hypervolume, ament2023unexpected}. (iii) \textbf{Information theoretic-based methods} aim to maximize information gain about the PF by selecting candidate points. PESMO~\cite{hernandez2016predictive} reduces uncertainty in the posterior over the Pareto optimal input set. Some extensions~\cite{belakaria2019max, suzuki2020multi, tu2022joint, hvarfner2022joint} refine this by considering output space. USeMO\cite{belakaria2020uncertainty} offers an alternative uncertainty measure based on hyper-rectangle volumes.

The solutions obtained by the aforementioned methods provide discrete approximations of the Pareto set. Lin et al.~\cite{lin2022pslmobo} propose learning a continuous approximation of the Pareto set in MOBO using a hypernetwork (introduced in Section \ref{sec:hypernetwork}). Subsequent improvements have been proposed to leverage diffusion models \cite{li2024expensive} and Stein variational gradient descent \cite{nguyen2024improving} for efficient solution generation, as well as incorporating data augmentation for Pareto set model building~\cite{lu2024you}.

\subsection{Computer Vision}
\label{sec:cv}
The most representative application of MOO in computer vision is multi-task dense prediction \cite{vandenhende2021multi}, which aims to train a model for simultaneously dealing with multiple dense prediction tasks (such as semantic segmentation, monocular depth estimation, and surface normal estimation) and has been successfully applied in autonomous driving \cite{ishihara2021multi,liang2023multi}. Since the encoder in computer vision usually contains large number of parameters, sharing the encoder across different tasks can significantly reduce the computational cost but may cause conflicts among tasks, leading to a performance drop in some of the tasks. Hence, many methods are proposed to address the issue from the perspective of MOO, which use scalarization to balance multiple losses \cite{xu2022mtformer,ye2022inverted,lin2024mtmamba, lin2024mtmambaplus,li2024universal}. A comprehensive survey on multi-task dense prediction is in \cite{vandenhende2021multi}.
Besides the dense prediction task, MOO is also useful in other computer vision tasks such as point cloud \cite{xie2023poly,xie2024co, wang2024one}, medical image denoising~\cite{kyung2024generative}, and pose estimation~\cite{ye2024lpformer}.

\subsection{Neural Architecture Search}
\label{sec:nas}
Neural Architecture Search (NAS), which aims to design the architecture of neural networks automatically, has gained significant interest recently \cite{ren2021comprehensive}. Due to the complex application scenarios in the real world, recent works consider multiple objectives beyond just accuracy. For example, to search an efficient architecture for deployment on platforms with limited resources, many studies incorporate resource-constraint objectives such as the parameter size, FLOPs, and latency. These works are mainly based on gradient-based NAS methods (like DARTS \cite{lsy19}) and formulate it as a multi-objective optimization problem. Among them, \cite{wu2019fbnet,CaiZH19, wu2021trilevel,wang2021attentivenas,yue2022effective} employ scalarization to identify a single solution, where the task and efficiency objectives are combined with fixed weights. However, altering the weights of the objectives needs a complete rerun of the search, which is computationally intensive. Therefore, Sukthanker et al. \cite{sukthanker2024multi} propose to learn a mapping from a preference vector to an architecture using a hypernetwork, enabling to provide the entire PS without the need for a new search.

\subsection{Recommendation Systems}
\label{sec:recomend}
Beyond just focusing on accuracy, recommendation systems often incorporate additional quality metrics such as novelty, diversity, serendipity, popularity, and others~\cite{zheng2022survey}. These factors naturally frame the problem as a multi-objective optimization optimization. Traditionally, many methods have employed scalarization techniques to assign weights to the different objectives~\cite{rsls1,rsls2,rstch}. In recent developments, gradient-balancing approaches are applied to address multi-objective optimization within recommendation systems to learn a single recommendation model~\cite{recommender1, recommender2, recommender3, recommender4}. Additionally, some research efforts aim to identify finite Pareto sets~\cite{rspareto0} and infinite Pareto sets~\cite{rspareto1, rspareto2}, enabling the provision of personalized recommender models that cater to varying user preferences.

\subsection{Large Language Models (LLMs)}
\label{sec:llm}
Recently, there has been a growing trend of incorporating multi-objective optimization into the training of large language models (LLMs), including multi-task fine-tuning and multi-objective alignment.

Due to the powerful transferability of LLMs, users can fine-tune LLMs to specific downstream tasks or scenarios. This approach separates fine-tuning on each task, causing extensive costs in training and difficulties in deployment. MFTCoder~\cite{liu2024mftcoder} proposes a multi-task fine-tuning framework that enhances the coding capabilities of LLMs by addressing data imbalance, varying difficulty levels, and inconsistent convergence speeds across tasks. CoBa~\cite{gong2024coba} studies multi-task fine-tuning for LLMs and balances task convergence with minimal computational overhead.

Aligning with multi-dimensional human preferences (such as helpfulness, harmfulness, humor, and conciseness) is essential for customizing responses to users' needs, as users typically have diverse preferences for different aspects. MORLHF~\cite{wu2023fine} and MODPO~\cite{zhou2024beyond} train an LLM for every preference configuration by linearly combining multiple (implicit) reward models. To avoid retraining, some methods train multiple LLMs separately for each preference dimension and merge their parameters \cite{rame2023rewarded, jang2023personalized} or output logits \cite{shi2024decoding} to deal with different preference requirements. However, these methods need to train and store multiple LLMs, leading to huge computational and storage costs. To improve efficiency, several preference-aware methods are proposed to fine-tune a single LLM for varying preferences by incorporating the relevant coefficients into the input prompts \cite{wang2024arithmetic, guo2024controllable, yangrewards} or model parameters \cite{panacea,lin2025parm}.

\subsection{Miscellaneous}
\label{sec:miscellaneous}
In addition to the applications mentioned above, multi-objective optimization has been applied to a variety of other deep learning scenarios, including meta-learning~\cite{yu2023enhancing,wang2021bridging}, federated learning~\cite{hu2022federated, askin2024federated, kang2024optimizing}, long-tailed learning~\cite{li2024long, zhao2024two, zhoupareto}, continual learning~\cite{paretoCL}, diffusion models~\cite{hang2023efficient, go2024addressing, xu2024diffusion, yao2024proud}, neural combinatorial optimization~\cite{li2020deep, lin2022pareto_combinatorial, wang2023multiobjective, chen2023neural, chen2023efficient}, GFlowNets~\cite{jain2023multi, zhu2023sample}, and physics informed
neural networks (PINNs) \cite{hwang2024dual,liu2024config}. These applications highlight the versatility of multi-objective optimization in addressing diverse challenges within deep learning.

\begin{table}[!t]
\caption{Summary of benchmark datasets in multi-objective deep learning.}
\vspace{-0.1in}
\label{tab:dataset}
\centering
% \resizebox{0.6\linewidth}{!}{
\begin{tabular}{lccc}
\toprule
Dataset & Description & \#Objectives & \#Samples \\
\midrule
\href{https://cs.nyu.edu/~fergus/datasets/nyu_depth_v2.html}{NYUv2} \cite{silberman2012indoor} & indoor scene understanding & $3$ & $1,449$\\
\href{http://taskonomy.stanford.edu/}{Taskonomy} \cite{zamir2018taskonomy} & indoor scene understanding & $26$ & $\approx4$M\\
\href{https://www.cityscapes-dataset.com/}{Cityscapes} \cite{CordtsORREBFRS16} & urban scene understanding & $2$ & $3,475$\\
\href{https://pytorch-geometric.readthedocs.io/en/2.6.1/generated/torch_geometric.datasets.QM9.html}{QM9} \cite{ramakrishnan2014quantum} & molecular property prediction & $11$ & $130$K \\
\href{https://www.cc.gatech.edu/~judy/domainadapt/\#datasets_code}{Office-31} \cite{saenko2010adapting} & image classification & $3$ & $4,110$ \\
\href{https://www.hemanthdv.org/officeHomeDataset.html}{Office-Home} \cite{venkateswara2017deep} & image classification & $4$ & $15,500$ \\
\href{https://mmlab.ie.cuhk.edu.hk/projects/CelebA.html}{CelebA} \cite{liu2015deep} & image classification & $40$ & $\approx202$K \\
\href{https://www.cs.toronto.edu/~kriz/cifar.html}{CIFAR-100} \cite{krizhevsky2009learning} & image classification & $20$ & $60$K\\
\href{https://github.com/Farama-Foundation/Metaworld}{MT10/MT50} \cite{yu2020meta} & reinforcement learning & $10/50$ & - \\
\href{https://github.com/google-research/xtreme?tab=readme-ov-file\#download-the-data}{XTREME} \cite{hu2020xtreme} & multilingual learning & $9$ & $\approx597$K \\
\href{https://huggingface.co/datasets/PKU-Alignment/PKU-SafeRLHF}{PKU-SafeRLHF}~\cite{ji2024pku} & two-dimensional preference data & $2$ & $\approx82$K \\ 
\href{https://huggingface.co/datasets/openbmb/UltraFeedback}{UltraFeedBack}~\cite{cui2023ultrafeedback} & multi-dimensional preference data & $4$ & $\approx64$K \\ 
\href{https://huggingface.co/datasets/nvidia/HelpSteer2}{HelpSteer2}~\cite{wang2024helpsteer2} & multi-attributes preference data & $5$ & $\approx21$K\\
\bottomrule
\end{tabular}
\end{table}

\section{Resources}
\label{sec:resource}
In this section, we introduce widely-used benchmark datasets and open-source libraries for gradient-based MOO in deep learning. 
\subsection{Datasets}
Below, we provide details on the benchmark datasets, which are summarized in Table \ref{tab:dataset}:
\begin{itemize}
	\item \textbf{NYUv2} dataset \cite{silberman2012indoor} is for indoor scene understanding. It has three tasks (i.e., semantic segmentation, depth estimation, and surface normal prediction) with $795$ training and $654$ testing samples.
	\item \textbf{Taskonomy} dataset \cite{zamir2018taskonomy} is obtained from 3D scans of about $600$ buildings and contains $26$ tasks (such as semantic segmentation, depth estimation, surface normal prediction, keypoint detection, and edge detection) with about $4$ million samples. 
	\item \textbf{Cityscapes} dataset \cite{CordtsORREBFRS16} is for urban scene understanding. It has two tasks (i.e., semantic segmentation and depth estimation) with $2,975$ training and $500$ testing samples.
	\item \textbf{QM9} dataset
	\cite{ramakrishnan2014quantum} is for molecular property prediction with $11$ tasks. Each task performs regression on one property. It contains $130,000$ samples.
	\item \textbf{Office-31} dataset \cite{saenko2010adapting} contains $4,110$ images from three domains (tasks): Amazon, DSLR, and Webcam. Each task has 31 classes.
	\item \textbf{Office-Home} dataset \cite{venkateswara2017deep} contains $15,500$ images from four domains (tasks): artistic images, clipart, product images, and real-world images. Each task has $65$ object categories collected under office and home settings.
	\item \textbf{CelebA}~\cite{liu2015deep} is a large-scale face attribute dataset comprising $202,599$ face images. Each image is annotated with $40$ binary attributes, resulting in $40$ distinct binary classification tasks.
	\item \textbf{CIFAR-100} \cite{krizhevsky2009learning} is an image classification dataset with $100$ classes, containing $50$K training and $10$K testing images. These $100$ classes are divided into $20$ tasks, each of which is a $5$-class image classification problem. 
	\item \textbf{MT10} and \textbf{MT50} from Meta-World benchmark \cite{yu2020meta} are two multi-task reinforcement learning settings, containing $10$ and $50$ robot manipulation tasks, respectively.
	\item \textbf{XTREME} benchmark \cite{hu2020xtreme} aims to evaluate the cross-lingual generalization abilities of multilingual representations. It contains $9$ tasks in $40$ languages, including $2$ classification tasks, $2$ structure prediction tasks, $3$ question-answering tasks, and $2$ sentence retrieval tasks.
	\item \textbf{PKU-SafeRLHF} dataset \cite{ji2024pku} focuses on safety alignment in LLMs, containing $82,118$ question-answering pairs with the annotations of helpfulness and harmlessness. 
	\item \textbf{UltraFeedBack} dataset~\cite{cui2023ultrafeedback} contains $63,967$ prompts, each corresponding to four responses from different LLMs. Each response has $4$ aspects of annotations, namely instruction-following, truthfulness, honesty, and helpfulness, generated by GPT-4.  
	\item \textbf{HelpSteer2} dataset \cite{wang2024helpsteer2} contains $21, 362$ samples. Each sample contains a prompt and a response with $5$ human-annotated attributes (i.e., helpfulness, correctness, coherence, complexity, and verbosity), each ranging between $0$ and $4$ where higher means better for each attribute.
\end{itemize}

\subsection{Libraries}
Two popular libraries, LibMTL\footnote{\url{https://github.com/median-research-group/LibMTL}} \cite{lin2023libmtl} and LibMOON\footnote{\url{https://github.com/xzhang2523/libmoon}} \cite{zhang2024libmoon}, provide unified environments for implementing and fairly evaluating MOO methods. Both are built on PyTorch \cite{pytorch} and feature modular designs, enabling flexible development of new methods or application of existing ones to new scenarios.
LibMTL \cite{lin2023libmtl} mainly focuses on finding a single Pareto-optimal solution and supports $24$ methods introduced in Section \ref{sec:single_model} and $6$ benchmark datasets.  
LibMOON \cite{zhang2024libmoon} mainly focuses on exploring the whole Pareto set. It includes over 20 methods for obtaining a finite set of solutions (introduced in Section \ref{sec:pareto_set}) or infinite set of solutions (introduced in Section \ref{sec:pareto_infinite}).

\section{Challenges and Future Directions}
\label{sec:future}
Despite significant progress in applying MOO in deep learning, both in learning a single model and in learning a Pareto set of models, several challenges remain. 

\vspace{0.1in}
\noindent \textbf{Theoretical Understanding.}
While practical methods for MOO in deep learning have seen significant progress, their theoretical foundations remain relatively underexplored.
Most existing research focuses on analyzing the convergence of MOO algorithms to stationary points, while generalization error, which is critical for evaluating performance on unseen data, has received less attention. For instance, Chen et al.~\cite{chen2024three}, as discussed in Section \ref{sec:theroy_generalization}, provide the algorithm-dependent generalization analysis. Extending this to a broader range of algorithms could offer deeper insights into how different techniques affect generalization.
For Pareto set learning algorithms, there is limited theoretical understanding of how effectively existing algorithms can approximate Pareto sets. Zhang et al.~\cite{zhang2023hypervolume} established a generalization bound for HV-based Pareto set learning. However, the effect of various network architectures (Section~\ref{sec:network_structures}) on approximating Pareto sets remain unclear. Additionally, it is uncertain how effective current training strategies (Section~\ref{sec:training_strategy}) are in approximating the Pareto set.

\vspace{0.1in}
\noindent \textbf{Reducing Gradient Balancing Costs.}
While gradient balancing methods in Section~\ref{sec:grad_balancing} often show strong performance in various applications, they come with significant computational overhead. Despite the introduction of some practical speedup strategies (discussed in Section \ref{sec:speedup}), existing approaches remain insufficiently efficient. A deeper understanding of the optimization differences between gradient balancing, linear scalarization, and loss balancing is crucial. This insight could facilitate the integration of gradient balancing with linear scalarization and loss balancing, reducing computational overhead significantly and enabling its application in large-scale training scenarios.

\vspace{0.1in}
\noindent \textbf{Dealing with Large Number of Objectives.}
Some real-world problems involve handling a large number of objectives, which poses significant challenges for current Pareto set learning algorithms in Section~\ref{sec:pareto_infinite}. As the number of objectives grows, the preference vector space expands exponentially, making it challenging for existing random sampling-based techniques to effectively learn the mapping between preference vectors and solutions. Developing efficient sampling strategies for high-dimensional preference spaces remains a important research problem. Additionally, exploring methods to reduce or merge objectives based on their properties can be a promising approach to minimize the total number of objectives.

\vspace{0.1in}
\noindent \textbf{Distributed Training.}
Current MOO algorithms in deep learning are limited to single GPUs or machines, but scaling them to multi-GPU and distributed environments is increasingly important as models and datasets grow. This scaling introduces unique challenges compared to single-objective optimization. Gradient-balancing algorithms require efficient gradient distribution and communication across GPUs. Learning a set of Pareto-optimal solutions involves synchronizing multiple models across nodes with minimal overhead. Additionally, when data for different objectives is distributed across devices and cannot be shared (e.g., due to privacy constraints), collaborative computation of solutions or Pareto sets without direct data sharing becomes an important challenge.

\vspace{0.1in}
\noindent \textbf{Advancements in Large Language Models (LLMs).} 
As discussed in Section~\ref{sec:llm}, current research in MOO for LLMs largely focuses on the RLHF stage. Expanding MOO techniques to other stages in the LLM lifecycle, such as pre-training and inference, is a valuable direction for future research. Addressing these challenges can lead to models that are better aligned with user needs across their entire development processes. Additionally, user preferences are currently modeled as a preference vector. However, this representation may oversimplify more complex user preferences on LLMs. Exploring advanced methods to capture and represent these intricate preferences can enhance the effectiveness and personalization of LLMs.

\vspace{0.1in}
\noindent \textbf{Application in More Deep Learning Scenarios.}
While MOO has already been utilized in various deep learning scenarios, as highlighted in Section~\ref{sec:application}, there remain numerous unexplored areas within this field. Multi-objective characteristics are inherently present in most deep learning problems, as models are typically developed or evaluated based on multiple criteria. Consequently, trade-offs often arise naturally. Leveraging MOO methods to effectively address these trade-offs presents an opportunity for further exploration.

\section{Conclusion} \label{sec:conclusion}
In this paper, we provide the first comprehensive review of gradient-based multi-objective deep learning, a field of growing importance as models are required to balance multiple, often conflicting, objectives. We have systematically surveyed the spectrum of algorithms, covering methods for finding a single balanced model, a finite set of models, and an entire infinite Pareto set of models.

The review also delves into the theoretical foundations of these methods, summarizing key results in convergence while highlighting the need for a deeper understanding of generalization. The practical significance of MOO is demonstrated across diverse applications, including its emerging role in Large Language Models (LLMs). By unifying these approaches and identifying key challenges, this paper serves as a foundational resource to guide and inspire future advancements in this critical and rapidly evolving domain.

%%
%% The acknowledgments section is defined using the "acks" environment
%% (and NOT an unnumbered section). This ensures the proper
%% identification of the section in the article metadata, and the
%% consistent spelling of the heading.
% \begin{acks}
% \end{acks}

%%
%% The next two lines define the bibliography style to be used, and
%% the bibliography file.
\bibliographystyle{ACM-Reference-Format}
\bibliography{ref}

%%
%% If your work has an appendix, this is the place to put it.
% \appendix

\end{document}